\documentclass[journal]{IEEEtran}
\IEEEoverridecommandlockouts
\usepackage{epsfig,amsmath,algorithm,algpseudocode,amssymb,epsf,cite,subfigure,scalefnt,graphicx,epstopdf,setspace,color,diagbox,multirow,makecell}
\usepackage{threeparttable}


\def\cA{{\mathcal{A}}}   
\def\cE{{\mathcal{E}}}   
\def\cI{{\mathcal{I}}}  \def\cK{{\mathcal{K}}} \def\cL{{\mathcal{L}}}
 \def\cN{{\mathcal{N}}} \def\cO{{\mathcal{O}}} \def\cP{{\mathcal{P}}}
  \def\cS{{\mathcal{S}}} \def\cT{{\mathcal{T}}}
\def\cU{{\mathcal{U}}}

\def\ba{{\mathbf{a}}}    \def\be{{\mathbf{e}}}

\def\bp{{\mathbf{p}}} \def\bq{{\mathbf{q}}}   
\def\bu{{\mathbf{u}}} \def\bv{{\mathbf{v}}} \def\bw{{\mathbf{w}}} \def\bx{{\mathbf{x}}}

\def\bA{{\mathbf{A}}} \def\bB{{\mathbf{B}}} \def\bC{{\mathbf{C}}} \def\bD{{\mathbf{D}}} 
  \def\bH{{\mathbf{H}}} \def\bI{{\mathbf{I}}} 
 \def\bL{{\mathbf{L}}}  \def\bN{{\mathbf{N}}} 
\def\bP{{\mathbf{P}}} \def\bQ{{\mathbf{Q}}}  \def\bS{{\mathbf{S}}} 
\def\bU{{\mathbf{U}}} \def\bV{{\mathbf{V}}}  \def\bX{{\mathbf{X}}} \def\bY{{\mathbf{Y}}}

\def\argmin{\mathop{\mathrm{argmin}}}

\def\bzeros{\mathbf{0}}
\def\bones{\mathbf{1}}

     \def\d4{\!\!\!\!}

   \def\bLam{\mathbf{\Lambda}}
    
\def\bthe{\boldsymbol{\theta}}  \def\blam{\boldsymbol{\lambda}} \def\bpsi{\boldsymbol{\psi}} \def\bxi{\boldsymbol{\xi}}

  \def\R{{\mathbb{R}}}   \def\S{{\mathbb{S}}} \def\B{{\mathbb{B}}}
  \def\-{\! - \!}  \def\+{\! + \!}  \def\={\! = \!}  \def\>{\! > \!} \def\nn{\nonumber}

\newtheorem{proposition}{Proposition}

\newtheorem{lemma}{Lemma}

\newtheorem{remark}{Remark}

\newcommand{\bef}{\begin{figure}}
\newcommand{\eef}{\end{figure}}
\newcommand{\beq}{\begin{eqnarray}}
\newcommand{\eeq}{\end{eqnarray}}

\newenvironment{proof}[1][Proof]{\begin{trivlist}
\item[\hskip \labelsep {\bfseries #1}]}{\end{trivlist}}
\newenvironment{example}[1][Example]{\begin{trivlist}
\item[\hskip \labelsep {\bfseries #1}]}{\end{trivlist}}

\newcommand{\qed}{\nobreak \ifvmode \relax \else
\ifdim\lastskip<1.5em \hskip-\lastskip \hskip1.5em plus0em
minus0.5em \fi \nobreak \vrule height0.5em width0.5em
depth0.25em\fi}

\title{Dual Optimization for Kolmogorov Model Learning Using Enhanced Gradient Descent}
\author{Qiyou~Duan, Hadi~Ghauch,~\IEEEmembership{Member,~IEEE,}, and Taejoon~Kim,~\IEEEmembership{Senior Member,~IEEE}	
\thanks{ Q. Duan is with China Southern Power Grid Technology Co. Ltd., CSG, Guangzhou, China (e-mail: qyduan.ee@my.cityu.edu.hk).

         H. Ghauch is with the Department of COMELEC, Telecom-ParisTech, Paris, France (e-mail: hadi.ghauch@telecom-paristech.fr).

         T. Kim is with the Department of Electrical Engineering and Computer Science, The University of Kansas, Lawrence, KS 66045 USA (e-mail: taejoonkim@ku.edu). 
         
         The work of Taejoon Kim was supported in part by the National Science Foundation (NSF) under Grant CNS1955561, and in part by the Office of Naval Research (ONR) under Grant N00014-21-1-2472.}
}

\begin{document}

\maketitle

\begin{abstract}
Data representation techniques have made a substantial contribution to advancing data processing and machine learning (ML). Improving predictive power was the focus of previous representation techniques, which unfortunately perform rather poorly on the interpretability in terms of extracting underlying insights of the data. Recently, the Kolmogorov model (KM) was studied, which is an interpretable and predictable representation approach to learning the underlying probabilistic structure of a set of random variables. The existing KM learning algorithms using semi-definite relaxation with randomization (SDRwR) or discrete monotonic optimization (DMO) have, however, limited utility to big data applications because they do not scale well computationally.
In this paper, we propose a computationally scalable KM learning algorithm, based on the regularized dual optimization combined with enhanced gradient descent (GD) method.
To make our method more scalable to large-dimensional problems, we propose two acceleration schemes, namely, the eigenvalue decomposition (EVD) elimination strategy and an approximate EVD algorithm. Furthermore, a thresholding technique by exploiting the error bound analysis and leveraging the normalized Minkowski $\ell_1$-norm, is provided for the selection of the number of iterations of the approximate EVD algorithm.
When applied to big data applications, it is demonstrated that the proposed method can achieve compatible training/prediction performance with significantly reduced computational complexity; roughly two orders of magnitude improvement in terms of the time overhead, compared to the existing KM learning algorithms. Furthermore, it is shown that the accuracy of logical relation mining for interpretability by using the proposed KM learning algorithm exceeds $80\%$.
\end{abstract}


\section{Introduction} \label{Sec.1} % Sec.1 Introduction
The digital era, influencing and reshaping the behaviors, performances, and standards, etc., of societies, communities, and individuals, has presented a big challenge for the conventional mode of data processing. Data consisting of numbers, words, and measurements becomes available in such huge volume, high velocity, and wide variety that it ends up outpacing human-oriented computing. It is urgent to explore the intelligent tools necessary for processing the staggering amount of data. Machine learning (ML), dedicated to providing insights into patterns in big data and extracting pieces of information hidden inside, arises and has been used in a wide variety of applications, such as computer vision \cite{FZhang18}, telecommunication \cite{Saad19}, and recommendation systems \cite{Faryal15, Narayanan16, Yi17,Nawrocka18}. Nevertheless, traditional ML algorithms become computationally inefficient and fail to scale up well as the dimension of data grows. A major issue that remains to be addressed is to find effective ML algorithms that perform well on both predictability and interpretability as well as are capable of tackling large-dimensional data with low complexity.

\subsection{Related Work}
Data representation, providing driving forces to the advancing ML-based techniques, has lately attracted a great deal of interest because it transforms large-dimensional data into low-dimensional alternatives by capturing their key features and make them amenable for processing, prediction, and analysis. The gamut of data representation techniques including matrix factorization (MF) \cite{Koren09,Ren18}, singular value decomposition (SVD)-based models \cite{Koren08,Cao15}, nonnegative models (NNM) \cite{Stark16}, and deep neural networks \cite{LeCun15} have been shown to perform well in terms of predictive power (the capability of predicting the outcome of random variables that are outside the training set). Unfortunately, these techniques perform rather poorly on the interpretability (the capability of extracting additional information or insights that are hidden inside the data) because on the one hand, they are not developed to directly model the outcome of random variables; on the other hand, they fall under the black-box category which lacks transparency and accountability of predictive models \cite{Rudin19}. Recently, a Kolmogorov model (KM) that directly represents a binary random variable as a superposition of elementary events in probability space was proposed \cite{Ghauch18}; KM models the outcome of a binary random variable as an inner product between two structured vectors, one probability mass function vector and one binary indicator vector. This inner product structure exactly represents an actual probability. Carefully examining association rules between two binary indicator vectors grants the interpretability of KM that establishes mathematically logical/causal relations between different random variables.

Previously, the KM learning was formulated as a coupled combinatorial optimization problem \cite{Ghauch18} by decomposing it into two subproblems: i) linearly-constrained quadratic program (LCQP) and ii) binary quadratic program (BQP), which can be alternatively solved by utilizing block coordinate descent (BCD). An elegant, low-complexity Frank-Wolfe (FW) algorithm \cite{Jaggi13} was used to optimally solve the LCQP by exploiting the unit probability simplex structure. Whereas, it is known to be unpromising to find algorithms to exactly solve the BQP problems in polynomial time. To get around this challenge, relaxation methods for linear \cite{Wainwright05, Kolmogorov06}, quadratic \cite{Ravikumar06}, second-order cone \cite{Kumar06, Ghaddar11}, and semi-definite programming (SDP) \cite{Kim03, Jordan04}, were proffered to produce a feasible solution close to the optimal solution of the original problem. Among these relaxation methods, the semi-definite relaxation (SDR) has been shown to have a tighter approximation bound than that of others \cite{Vandenberghe1996,Wolkowicz00}. Thus, an SDR with randomization (SDRwR) method \cite{Kisialiou10} was employed to optimally solve the BQP of the KM learning in an asymptotic sense \cite{Ghauch18}. To address the high-complexity issue due to the reliance on the interior point methods, a branch-reduce-and-bound (BRB) algorithm based on discrete monotonic optimization (DMO) \cite{Kim15,Duan20} was proposed. However, the DMO approach only shows its efficacy in a low-dimensional setting and starts to collapse as the dimension increases. In short, the existing KM methods \cite{Ghauch18,Duan20} suffer from a similar drawback, namely, being unscalable. Unfortunately, the latter limitation hampers the application of them to large-scale datasets, for instance, the MovieLens 1 million (ML1M) dataset\footnote{https://grouplens.org/datasets/movielens/1m/}. It is thus crucial to explore low-complexity and scalable methods for KM learning.

Duality often arises in linear/nonlinear optimization models in a wide variety of applications such as communication networks \cite{Xiong21}, economic markets \cite{Peterson1970}, and structural design \cite{Larsson1993}. Simultaneously, the dual problem possesses some good mathematical, geometric, or computational structures that can be exploited to provide an alternative way of handling the intricate primal problems by using iterative methods, such as the first-order gradient descent (GD) \cite{Boyd04, Kim16} and quasi-Newton method \cite{Broyden1967, Nocedal06}. It is for this reason that the first-order iterative methods are widely used when optimizing/training large-scale data representations (e.g., deep neural networks) and machine learning algorithms. We are motivated by these iterative first-order methods to effectively resolve the combinatorial challenge of KM learning.

\subsection{Overview of Methodologies and Contributions}
We present a computationally scalable approach to the KM learning problem by proposing an enhanced GD algorithm and an approximate eigenvalue decomposition (EVD) with thresholding scheme based on dual optimization. Our main contributions are listed below.
\begin{itemize}
  \item We provide a reformulation of the BQP subproblem of KM learning to a regularized dual optimization problem that ensures strong duality and is amenable to be solved by simple GD. Compared to the existing SDRwR \cite{Ghauch18} and DMO \cite{Duan20}, the proposed dual optimization method proffers a more efficient and scalable solution to KM learning. This algorithmic approach is ideally suited to the KM learning, but is not limited thereto, and can be applied to any realistic problem involving BQP.
  \item Motivated by the fact that EVD is required at each iteration of GD, which introduces a computational bottleneck when applied to big data, an enhanced GD that eliminates the EVD computation when it is feasible is proposed to accelerate the computational speed. When the elimination is infeasible and EVD must be computed, we explore an approximate EVD based on the Lanczos method \cite{Golub1996} by taking account of the fact that computing exact, entire EVD is usually unnecessary. We focus on analyzing the approximation error of the approximate EVD. A tractable thresholding scheme is then proposed to determine the number of iterations of the approximate EVD by exploiting the structure of the upper bound on the approximation error and utilizing the normalized Minkowski $\ell_1$-norm.
  \item Extensive numerical simulation results are presented to demonstrate the efficacy of the proposed KM learning algorithm. When applied to large-scale datasets (e.g., ML1M dataset), it is shown that the proposed method can achieve comparable training and prediction performance with significantly reduced computational cost of more than two orders of magnitude, compared to the existing KM learning algorithms. Finally, the interpretability of the proposed method is validated by exploiting the mathematically logical relations. We show that the accuracy of logical relation mining by using the proposed method exceeds $80\%$.
\end{itemize}

\emph{Notation:} A bold lowercase letter $\ba$ is a vector and a bold capital letter $\bA$ is a matrix. $A(i,j)$, $\bA(:,j)$, $\text{trace}(\bA)$, $\text{diag}(\bA)$, $\text{rank}(\bA)$, $\lambda_{\text{max}}(\bA)$, and $\sigma_{\text{max}}(\bA)$ denote the $(i,j)$th entry, $j$th column, trace, main diagonal elements, rank, largest eigenvalue, and largest singular value of $\bA$, respectively. $a(i)$ is the $i$th entry of $\ba$, $\ba(m:n)\triangleq[a(m),\cdots,a(n)]^T$, and $\text{diag}(\ba)$ is a diagonal matrix with $\ba$ on its main diagonal. $\langle \bX, \bY \rangle$ is the Frobenius inner product of two matrices $\bX$ and $\bY$, i.e., $\langle \bX, \bY \rangle=\text{trace}(\bX^T\bY)$. $\bX\succeq \bzeros$ indicates that the matrix $\bX$ is positive semi-definite (PSD). $\be_i$ is the $i$th column of the identity matrix of appropriate size. $\bones$ and $\bzeros$ denote the all-one and all-zero vectors, respectively. $\S^{N\times N}$, $\R_+^N$, and $\B^N$ denote the $N\times N$ symmetric matrix space, nonnegative real-valued $N\times 1$ vector space, and $N\times 1$ binary vector space with each entry chosen from $\{0,1\}$, respectively. For $\bS\in\S^{N\times N}$, $\boldsymbol{\lambda}(\bS)\triangleq [\lambda_1(\bS), \lambda_2(\bS), \cdots, \lambda_N(\bS)]^T\in\R^{N\times 1}$ where $\lambda_n(\bS)$ is the $n$th eigenvalue of $\bS$, $n=1,\ldots,N$. $\text{supp}(\ba)\triangleq \{i|a_i\neq 0, i\in\{1,\ldots,N\}\}$ is the support set of $\ba\in\R^N$ and $|\cA|$ denotes the cardinality of a set $\cA$. Finally, $\cE_1\Rightarrow \cE_2$ indicates that one outcome $\cE_1$ completely implies another one $\cE_2$.

\section{System Model and Preliminaries} \label{Sec.2} % Sec.2 System Model and Previous Work

In this section, we briefly discuss the concept of KM and its learning framework.

\subsection{Preliminaries} \label{Sec.2.1}
We consider a double-index set of binary random variables $X_{u,i}\in\{0,1\}$, $\forall (u,i)\in \cS$, where $\cS\triangleq\{(u,i)|(u,i)\in \cU\times\cI\}$ ($\cU=\{1,\ldots,U\}$ and $\cI=\{1,\ldots,I\}$ are the index sets of $u$ and $i$, respectively) denotes the set of all index pairs. Thus, $X_{u,i}$ can represent any two-dimensional learning applications (involving matrices) such as movie recommendation systems \cite{Stark16}, DNA methylation for cancer detection \cite{Houseman12}, and beam alignment in multiple-antenna systems \cite{Hur13, Alkhateeb14}.
We let $\Pr(X_{u,i}=1)\in [0, 1]$ be the probability that the event $X_{u,i}=1$ occurs.
Since the random variable considered here is binary, the following holds $\Pr(X_{u,i}=1)+\Pr(X_{u,i}=0)=1$. Without loss of generality, we can focus on one outcome, for instance, $X_{u,i}=1$. Then, the $D$-dimensional KM of the random variable $X_{u,i}$ is given by
\beq \label{KM}
\Pr(X_{u,i}=1) = \bthe_u^T\bpsi_i,~\forall (u,i)\in\mathcal{S},
\eeq
where $\bthe_u \in \R_+^D$ is the \emph{probability mass function vector} and $\bpsi_i \in\B^D$ is the \emph{binary indicator vector}.
Specifically, $\bthe_u$ is in the unit probability simplex $\cP\triangleq\{\bp\in\R_+^D|\bones^T\bp=1\}$, i.e., $\bthe_u\in\cP$, and $\bpsi_i$ denotes the support set of $X_{u,i}$ (associated with the case when $X_{u,i}=1$). The KM in \eqref{KM} is built under a measurable probability space defined on $(\Omega, \cE)$ ($\Omega$ denotes the sample space and $\cE$ is the event space consisting of subsets of $\Omega$) and satisfies the following conditions: i) $\Pr(E)\geq 0$, $\forall E\in\cE$ (nonnegativity), ii) $\Pr(\Omega) = 1$ (normalization), and iii) $\Pr(\cup_{i=1}^\infty E_i) = \sum_{i=1}^{\infty}\Pr(E_i)$ for the disjoint events $E_i\in\cE$, $\forall i$ (countable additivity) \cite{Gray09}. By \eqref{KM}, $X_{u,i}$ is modeled as stochastic mixtures of $D$ Kolmogorov elementary events. In addition, note that $\Pr(X_{u,i}=0) = \bthe_u^T(\bones-\bpsi_i)$.

\subsection{KM Learning} \label{Sec.2.2}
Assume that the empirical probability of $X_{u,i}=1$, denoted by $p_{u,i}$, is available from the training set $\cK\triangleq\{(u,i)|u\in\cU_{\cK}\subseteq\cU, i\in\cI_{\cK}\subseteq\cI\}\subseteq \cS$. Obtaining the empirical probabilities $\{p_{u,i}\}$ for the training set depends on the application and context in practical systems; we will illustrate an example for recommendation systems at the end of this section. The KM learning involves training, prediction, and interpretation as described below.
\subsubsection{Training}
The KM training proceeds to optimize $\{\bthe_u\}$ and $\{\bpsi_i\}$ by solving the $\ell_2$-norm minimization problem:
\beq \label{KM problem formulation}
\begin{aligned}
\{\boldsymbol{\theta}_u^\star\}, \{\boldsymbol{\psi}_i^\star\} &= \argmin_{\{\boldsymbol{\theta}_u\},\{\boldsymbol{\psi}_i\}} \sum_{(u,i)\in\cK} ( \boldsymbol{\theta}_u^T\boldsymbol{\psi}_i - p_{u,i} )^2 \\
   & \text{s.t.}\ \boldsymbol{\theta}_u \in \cP, \ \boldsymbol{\psi}_i\in \B^D, \forall (u,i) \in \cK
\end{aligned}.
\eeq
To deal with the coupled combinatorial nature of \eqref{KM problem formulation}, a BCD method \cite{Ghauch18,Chan19} was proposed by dividing the problem in \eqref{KM problem formulation} into two subproblems:
i) LCQP:
\beq \label{subproblem 1: LCQP}
\bthe_u^{(\tau+1)} \!\!=\! \argmin_{\boldsymbol{\theta}_u \in \cP} \boldsymbol{\theta}_u^T\bQ_u^{(\tau)} \boldsymbol{\theta}_u - 2\boldsymbol{\theta}_u^T\bw_u^{(\tau)} + \varrho_u, \ \forall u \in\cU_{\cK},
\eeq
where $\bQ_u^{(\tau)}\triangleq \sum_{i \in \cI_{u}} \boldsymbol{\psi}_i^{(\tau)}{\boldsymbol{\psi}_i^{(\tau)}}^T$, $\bw_u^{(\tau)} \triangleq \sum_{i \in \cI_{u}} \boldsymbol{\psi}_i^{(\tau)} p_{u,i}$, $\varrho_u \triangleq \sum_{i \in \cI_{u}} p_{u,i}^2$, $\cI_u\triangleq \{i|(u,i)\in \cK\}$, and $\tau$ is the index of BCD iterations,
and ii) BQP:
\beq \label{subproblem 2: BQP}
\!\bpsi_i^{(\tau+1)} \!\!=\! \argmin_{\boldsymbol{\psi}_i\in \B^D} \boldsymbol{\psi}_i^T\bS_i^{(\tau+1)}\boldsymbol{\psi}_i \!-\! 2\boldsymbol{\psi}_i^T\bv_i^{(\tau+1)} \!\!+\! \rho_i, \forall i \!\in\!\cI_{\cK},
\eeq
where $\bS_i^{(\tau+1)}\triangleq \sum_{u\in \cU_{i}} \boldsymbol{\theta}_u^{(\tau+1)}{\boldsymbol{\theta}_u^{(\tau+1)}}^T$, $\bv_i^{(\tau+1)}\triangleq \sum_{u\in \cU_{i}}$ $\boldsymbol{\theta}_u^{(\tau+1)} p_{u,i}$, $\rho_i \triangleq \sum_{u\in \cU_{i}} p_{u,i}^2$, and $\cU_{i}\triangleq \{u|(u,i)\in \cK\}$.
BCD has been successful in tackling coupled optimization problems in applications such as the transceiver design in wireless communications \cite{Gomadam11,Shi11,Ghauch16,Xiong21,Zhang18}. The coupling among $\{\bthe_u\}$ and $\{\boldsymbol{\psi}_i\}$ in (2) makes BCD an ideal method to alternatively handle the coupled optimization problem. It has been studied that the BCD method converges to a local minimum of the original problem in (2) if a unique minimizer is found for both blocks, $\{\bthe_u\}$ and $\{\boldsymbol{\psi}_i\}$ \cite{Tseng01,Ghauch18}.

By exploiting the fact that the optimization in \eqref{subproblem 1: LCQP} was carried out over the unit probability simplex $\cP$, a simple iterative FW algorithm \cite{Jaggi13} was employed to optimally solve \eqref{subproblem 1: LCQP}, while the SDRwR was employed to asymptotically solve the BQP in \eqref{subproblem 2: BQP} \cite{Kisialiou10}. It is also possible to solve \eqref{subproblem 2: BQP} directly without a relaxation and/or randomization, based on the DMO approach \cite{Duan20}. However, the DMO in \cite{Duan20} was shown only to be efficient when the dimension $D$ is small (e.g., $D\leq 8$); its computational cost blows up as $D$ increases (e.g., $D>20$).
\subsubsection{Prediction}
Similar to other supervised learning methods, the trained KM parameters $\{\boldsymbol{\theta}_u^\star\}, \{\boldsymbol{\psi}_i^\star\}$ are used to predict probabilities over a test set $\cT$ as
\beq \label{probabilities prediction}
\hat{p}_{u,i}\triangleq {\boldsymbol{\theta}_u^\star}^T\boldsymbol{\psi}_i^\star,\ \forall (u,i)\in \cT,
\eeq
where $\cT\cap \cK=\phi$ and $\cT\cup \cK =\cS$.
\subsubsection{Interpretation} \label{Section 2.2.3}
KM offers a distinct advantage, namely, the interpretability by drawing on fundamental insights into the mathematically logical relations among the data. For two random variables $X_{u,i}$ and $X_{u,j}$ taken from the training set $\cK$, i.e., $(u,i)\in\cK$ and $(u,j)\in\cK$, if the support sets of $\bpsi_i^\star$ and $\bpsi_j^\star$ satisfy $\text{supp}(\bpsi_j^\star)\subseteq \text{supp}(\bpsi_i^\star)$, then two logical relations between the outcomes of $X_{u,i}$ and $X_{u,j}$ can be inferred: the first outcome of $X_{u,i}$ implies the same one for $X_{u,j}$ while the second outcome of $X_{u,j}$ implies the second one for $X_{u,i}$, i.e.,
$X_{u,i} = 1 \Rightarrow  X_{u,j} = 1$ and $X_{u,j} = 0 \Rightarrow  X_{u,i} = 0$ \cite[Proposition 1]{Ghauch18}.
It is important to note that logical relations emerged from KM are based on the formalism of implications. Thus, they hold from a strictly mathematical perspective, and are general.

An implication of the introduced KM learning is illustrated by taking an example of movie recommendation systems as follows.
\begin{example}[Illustrative Example:] \label{illustrative example}
Suppose there are two users ($U=2$) who have rated two movie items ($I=2$). In this example, $X_{u,i}=1$ denotes the event that user $u$ likes the movie item $i$, $\forall u\in\{1,2\}, \forall
i\in\{1,2\}$. Then, $\Pr(X_{u,i}=1)$ denotes the probability that user $u$ likes item $i$ (conversely, $\Pr(X_{u,i}=0)$ denotes the probability that user $u$ dislikes item $i$). Suppose $D=4$ in \eqref{KM}. Then, the four elementary events can represent four different movie genres including i) Comedy, ii) Thriller, iii) Action, and iv) Drama. The empirical probability corresponding to $X_{u,i}=1$ can be obtained by
\beq \label{eq:empirical probability}
p_{u,i}\triangleq \frac{r_{u,i}}{r_{\text{max}}},
\eeq
where $r_{u,i}$ denotes the rating score that user $u$ has provided for item $i$ and $r_{\text{max}}$ is the maximum rating score.
In a 5-star rating system ($r_{\text{max}}=5$), we consider the following matrix as an example:
\beq \label{Toy example-empirical probabilities}
\Big[
  \begin{array}{cc}
    p_{1,1} & p_{1,2} \\
    p_{2,1} & p_{2,2} \\
  \end{array}
\Big] = \Big[
            \begin{array}{cc}
              0.8 & 0.4 \\
              * & 0.6 \\
            \end{array}
          \Big],
\eeq
where $p_{2,1}$ is unknown (as in the `*' entry) and $\{p_{1,1}, p_{1,2},$ $ p_{2,2}\}$ constitutes the training set of empirical probability where $\cK=\{(1,1),(1,2),(2,2)\}$.
By solving the KM learning problem in \eqref{KM problem formulation} for the empirical probabilities provided in \eqref{Toy example-empirical probabilities}, one can find the optimal model parameters, $\{\bthe_u^\star\}$ and $\{\bpsi_i^\star\}$ (an optimal solution to \eqref{KM problem formulation}), which is given by
\beq \label{Toy example: KM}
&&\bthe_1^\star = [0.4~ 0.2~ 0.1~ 0.3]^T, ~ \bthe_2^\star = [0.1~ 0.3~ 0.1~ 0.5]^T; \nn \\
&&\bpsi_1^\star = [1~ 0~ 1~ 1]^T, ~ \bpsi_2^\star = [0~ 0~ 1~ 1]^T. \nn
\eeq
\end{example}
Then, we can predict $p_{2,1}$ ($\cT=\{(2,1)\}$) by using the learned KM parameters $\bthe_2^\star$ and $\bpsi_1^\star$ as $\hat{p}_{2,1}={\bthe_2^\star}^T\bpsi_1^\star = 0.7$.
In this example, the following inclusion holds $\text{supp}(\bpsi_2^\star)\subset \text{supp}(\bpsi_1^\star)$. Thus, if a certain user (user 1 or 2) likes movie item 1, this logically implies that the user also likes movie item 2.

\begin{remark}
In contrast to the KM in \eqref{KM}, the state-of-the-art method, MF \cite{Koren09,Ren18}, considers an inner product of two arbitrary vectors without having implicit or desired structures in place. While NNM \cite{Stark16} has a similar structure as \eqref{KM}, the distinction is that NNM relaxes the binary constraints on $\bpsi_i$ to a nonnegative box, i.e., $\bpsi_i\in[\bzeros,\bones]$, and thus sacrifices the highly interpretable nature of KM. Unlike the existing data representation techniques, the KM can exactly represent the outcome of random variables in a Kolmogorov sense. As illustrated in Section \ref{Sec.5}, this in turn improves the prediction performance of the KM compared to other existing data representation techniques. Despite its predictability benefit, the existing KM learning methods \cite{Ghauch18,Duan20}, however, suffer from high computational complexity and a lack of scalability. In particular, the LCQP subproblem, which can be efficiently solved by the FW algorithm, has been well-investigated, while resolving the BQP introduces a major computational bottleneck. It is thus of great importance to study more efficient and fast KM learning algorithms that are readily applicable to large-scale problems.
\end{remark}

\section{Proposed Method} \label{Section 3}
To scale KM learning, we propose an efficient, first-order method to the BQP subproblem in \eqref{subproblem 2: BQP}.

\subsection{Dual Problem Formulation} \label{Section 3.1}
We transform the BQP subproblem in \eqref{subproblem 2: BQP} to a dual problem. To this end, we formulate an equivalent form to the BQP in \eqref{subproblem 2: BQP} as
\beq \label{reformulated BQP}
\min_{\bx\in\{+1,-1\}^D} \bx^T\bA_0\bx + \ba^T\bx,
\eeq
where $\rho_i$ in \eqref{subproblem 2: BQP} is ignored in \eqref{reformulated BQP}, $\bx = 2\bpsi_i - \bones \in\{+1,-1\}^D$, $\bA_0 = \frac{1}{4}\bS_i$, and $\ba = \frac{1}{2}\bS_i^T\bones - \bv_i$. For simplicity, the iteration index $\tau$ is omitted hereinafter. By introducing $\bX_0 = \bx\bx^T$ and $\bX= \Big[\begin{array}{cc}
                                                                                                      1 & \bx^T \\
                                                                                                      \bx & \bX_0 \\
                                                                                                      \end{array}\Big]\in\S^{(D+1)\times(D+1)}$, the problem in \eqref{reformulated BQP} can be rewritten as
\begin{subequations}\label{reformulated BQP+}
\begin{alignat}{2}
\min_{\bx,\bX_0} &\quad \langle\bX_0, \bA_0\rangle + \ba^T\bx, \label{reformulated BQP+:objective func} \\
\text{s.t.} &\quad \text{diag}(\bX_0) = \bones, \label{reformulated BQP+:diagonal constraint} \\
            &\quad \bX \succeq \bzeros, \label{reformulated BQP+:psd constraint} \\
            &\quad \text{rank}(\bX) = 1. \label{reformulated BQP+:rank constraint}
\end{alignat}
\end{subequations}
Solving \eqref{reformulated BQP+} directly is NP-hard due to the rank constraint in \eqref{reformulated BQP+:rank constraint}, thus we turn to convex relaxation methods. The SDR to \eqref{reformulated BQP+} can be expressed in a homogenized form with respect to $\bX$ as
\begin{subequations}\label{SDP relaxation}
\begin{alignat}{2}
  \min_{\bX} & \quad f(\bX)\triangleq \langle\bX, \bA\rangle, \label{SDP relaxation:objective function}  \\
  \text{s.t.} & \quad \langle\bB_i, \bX\rangle = 1, ~ i = 1,\ldots,D+1, \label{SDP relaxation:diagonal constraint} \\
              & \quad \bX \succeq \bzeros, \label{SDP relaxation: psd constraint}
\end{alignat}
\end{subequations}
where $\bA=\Big[\!\!\begin{array}{cc}
                      0\!\! & (1/2)\ba^T \\
                      (1/2)\ba\!\! & \bA_0 \\
                      \end{array}\!\!\Big]\in\S^{(D+1)\times(D+1)}$ and $\bB_i=[\bzeros_1~\cdots~\bzeros_{i-1}~\be_i~\bzeros_{i+1}~\cdots~\bzeros_{D+1}]\in\R^{(D+1)\times(D+1)}$.
Note that the diagonal constraint in \eqref{reformulated BQP+:diagonal constraint} has been equivalently transformed to $D+1$ equality constraints in \eqref{SDP relaxation:diagonal constraint}.
While the problem in \eqref{reformulated BQP+} is combinatorial due to the rank constraint, the relaxed problem in \eqref{SDP relaxation} is a convex SDP. Moreover, the relaxation is done by dropping the rank constraint.

We further formulate a regularized SDP formulation of \eqref{SDP relaxation} as
\beq \label{regularized SDP}
\min_{\bX} && f_\gamma(\bX)\triangleq \langle\bX, \bA\rangle + \frac{1}{2\gamma}\|\bX\|_F^2, \\
\text{s.t.} &&  \langle\bB_i, \bX\rangle = 1, ~ i = 1,\ldots,D+1, \nonumber \\
            &&  \bX \succeq \bzeros, \nonumber
\eeq
where $\gamma>0$ is a regularization parameter. With a Frobenius-norm term regularized, the strict convexity of \eqref{regularized SDP} is ensured, which in turn makes strong duality hold for the feasible dual problem of \eqref{regularized SDP}. In this work, we leverage this fact that the duality gap is zero for \eqref{regularized SDP} (a consequence of strong duality) to solve the dual problem. Using a larger regularization parameter $\gamma$ yields better quality of the solution to (10), but at the cost of slower convergence. In addition, the two problems in \eqref{SDP relaxation} and \eqref{regularized SDP} are equivalent as $\gamma\rightarrow\infty$. The choice of $\gamma$ will be further discussed in Section \ref{Sec.5.1}.

Given the regularized SDP formulation in \eqref{regularized SDP}, its dual problem and the gradient of the objective function are of interest.
\begin{lemma} \label{dual problem}
Suppose the problem in \eqref{regularized SDP} is feasible. Then, the dual problem of \eqref{regularized SDP} is given by
\beq \label{eq:dual}
\max_{\bu\in\R^{D+1}}\quad d_\gamma(\bu)\triangleq -\bu^T\bones - \frac{\gamma}{2}\|\Pi_+(\bC(\bu))\|_F^2,
\eeq
where $\bu\!\!\in\!\!\R^{D+1}$ is the vector of Lagrange \mbox{multipliers} associated with each of the $D+1$ equality constraints of \eqref{regularized SDP}, $\bC(\bu)\triangleq -\bA - \sum_{i=1}^{D+1}u_i\bB_i$, and $\Pi_+(\bC(\bu))\triangleq \sum_{i=1}^{D+1}\max(0,\lambda_i(\bC(\bu)))\bp_i\bp_i^T$, in which $\lambda_i(\bC(\bu))$ and $\bp_i$, $i=1,\ldots,D+1$, respectively, are the eigenvalues and corresponding eigenvectors of $\bC(\bu)$.
The gradient of $d_\gamma(\bu)$ with respect to $\bu$ is
\beq \label{eq:dual gradient}
\nabla_{\bu} d_\gamma(\bu)=-\bones + \gamma\Phi[\Pi_+(\bC(\bu))],
\eeq
where $\Phi[\Pi_+(\bC(\bu))]\triangleq [\langle\bB_1, \Pi_+(\bC(\bu))\rangle, \cdots, \langle\bB_{D+1},\Pi_+($ $\bC(\bu))\rangle]^T\in\R^{D+1}$.
\end{lemma}
\quad\textit{Proof:} See Appendix \ref{Proof of Lemma 1}.

\begin{algorithm}[t]
\caption{GD for Solving the Dual Problem in \eqref{convex transformation}} \label{Algorithm: Gradient Descent}
\begin{algorithmic}[1]
\Require
$\bA$, $\{\bB_i\}_{i=1}^{D+1}$, $D$, $\bu_0$, $\gamma$, $\epsilon$ (\emph{tolerance threshold value}), and $I_{\text{max}}$ (\emph{maximum number of iterations}).
\Ensure
$\bu^\star$.
\For{$i=0, 1, 2, \ldots, I_{\text{max}}$}
\State Calculate the gradient: $\nabla_{\bu_i} h_\gamma(\bu_i)$.
\State Compute the descent direction: $\Delta\bu_i = - \nabla_{\bu_i} h_\gamma(\bu_i)$.
\State Find a step size $t_i$ (via \emph{backtracking line search}), and $\bu_{i+1} = \bu_{i} + t_i\Delta\bu_i$.
\If {$\|t_i\Delta\bu_i\|_2\leq \epsilon$} terminate and
\Return $\bu^\star = \bu_{i+1}$.
\EndIf
\EndFor
\end{algorithmic}
\end{algorithm}

It is well known that $d_\gamma(\bu)$ in \eqref{eq:dual} is a strongly concave (piecewise linear) function, thereby making the Lagrange dual problem \eqref{eq:dual} a strongly convex problem having a unique global optimal solution \cite{Boyd04}. Furthermore, the special structure of $\bC(\bu)$ of Lemma \ref{dual problem}, i.e., being symmetric, allows us to propose computationally efficient and scalable KM learning algorithms which can be applied to handle large-scale datasets with low latency.

\subsection{Fast GD Methods For The Dual Problem} \label{Section 3.2}

\subsubsection{GD}
The dual problem in \eqref{eq:dual}, having a strongly concave function $d_\gamma(\bu)$, is equivalent to the following unconstrained convex minimization problem
\beq \label{convex transformation}
\min_{\bu\in\R^{D+1}}\quad h_\gamma(\bu)\triangleq \bu^T\bones + \frac{\gamma}{2}\|\Pi_+(\bC(\bu))\|_F^2,
\eeq
with the gradient being $\nabla_{\bu} h_\gamma(\bu)=\bones - \gamma\Phi[\Pi_+(\bC(\bu))]$. We first introduce a GD, which is detailed in Algorithm \ref{Algorithm: Gradient Descent}, to solve \eqref{convex transformation}. Note that, due to the fact that the dual problem in \eqref{convex transformation} is unconstrained, a simple GD method is proposed here: indeed, we would need a projected GD method if there is constraint included, for which the computational complexity would be much larger because of the projection at each iteration.

\begin{algorithm}[t]
\caption{Enhanced GD with EVD Elimination} \label{Algorithm: Enhanced Gradient Descent}
\begin{algorithmic}[1]
\Require
$-\bA=\bV\bLam\bV^T$, $\{\bB_i\}_{i=1}^{D+1}$, $D$, $\bu_0$~(\emph{with equal entries}), $\gamma$, $\epsilon$, and $I_{\text{max}}$.
\Ensure
$\bu^\star$.
\For{$i=0, 1, 2, \ldots, I_{\text{max}}$}
\State Calculate the gradient with EVD elimination:
\If {All $D+1$ elements of $\bu_i$ are the same}  \emph{\textbf{Phase I:}}
\State $\blam(\bC(\bu_i)) = \blam(-\bA) - \bu_i$ where $\blam(-\bA) = \text{diag}(\bLam)$.
\State Find the index set $\cI_{\lambda}\triangleq\{j|\lambda_j(\bC(\bu_i))>0, j=1,\ldots,D+1\}$.
   \If {$\cI_{\lambda}=\emptyset$} \emph{Phase I-A:}  $\nabla h_\gamma(\bu_i) = \bones$.
   \Else ~ \emph{Phase I-B:}
\State $\nabla h_\gamma(\bu_i)=\bones - \gamma\Phi[\Pi_+(\bC(\bu_i))]$, $\Pi_+(\bC(\bu_i))$ $=\sum_{j\in\cI_\lambda}\lambda_j(\bC(\bu_i))\bV(:,j)\bV(:,j)^T$.
   \EndIf
\Else ~ \emph{\textbf{Phase II:}}
   \If {$\lambda_{\text{max}}(-\bA) + \lambda_{\text{max}}(-\text{diag}(\bu_i))\leq 0$} \emph{Phase II-A:} $\nabla h_\gamma(\bu_i) = \bones$.
   \Else ~ \emph{Phase II-B:}
\State $\bC(\bu_i)=\bV_{\bC}\bLam_{\bC}\bV_{\bC}^T$, $\blam(\bC(\bu_i) = \text{diag}(\bLam_{\bC})$.
\State $\nabla h_\gamma(\bu_i)=\bones - \gamma\Phi[\Pi_+(\bC(\bu_i))]$, $\Pi_+(\bC(\bu_i))$ $=\sum_{j\in\cI_\lambda}\lambda_j(\bC(\bu_i))\bV_{\bC}(:,j)\bV_{\bC}(:,j)^T$.
   \EndIf
\EndIf
\State Compute the descent direction: $\Delta\bu_i = - \nabla h_\gamma(\bu_i)$.
\State Find a step size $t_i$ (via \emph{backtracking line search}), and $\bu_{i+1} = \bu_{i} + t_i\Delta\bu_i$.
\If {$\|t_i\Delta\bu_i\|_2\leq \epsilon$} terminate and
\Return $\bu^\star = \bu_{i+1}$.
\EndIf
\EndFor
\end{algorithmic}
\end{algorithm}

In Algorithm \ref{Algorithm: Gradient Descent}, only the gradient of $h_\gamma(\bu_i)$, i.e., $\nabla_{\bu_i}h_\gamma$ $(\bu_i)$, is required to determine the descent direction. It is therefore a more practical and cost-saving method compared to standard Newton methods which demand the calculation of second-order derivatives and the inverse of the Hessian matrix. Moreover, Algorithm \ref{Algorithm: Gradient Descent} does not rely on any approximation of the inverse of the Hessian matrix such as the quasi-Newton methods \cite{Wang17}. To find a step size in Step 4, we apply the backtracking line search method \cite{Bertsekas16}, which is based on the Armijo-Goldstein condition \cite{Armijo1966}. The algorithm is terminated when the pre-designed stopping criterion (for instance, $\|t_i\Delta\bu_i\|_2\leq \epsilon$ in Step 5, where $\epsilon>0$ is a predefined tolerance) is satisfied. Finally, the computational complexity of Algorithm \ref{Algorithm: Gradient Descent} is dominated by the EVD of a $(D+1)\times(D+1)$ matrix, needed to compute $\nabla_{\bu} h_\gamma(\bu)$ in Step 2, which is given as $\cO( (D+1)^3 )$.

\subsubsection{Enhanced GD}
In Algorithm \ref{Algorithm: Gradient Descent}, an EVD of $\bC(\bu_i)$ is required at each iteration to determine $\Pi_+(\bC(\bu_i))$ and $\nabla_{\bu_i} h_\gamma(\bu_i)$. However, it is difficult to employ EVD per iteration as they require high computational cost ($\cO(UI(D+1)^3)$) when large-scale datasets are involved (with very large $U$, $I$, and $D$). It is critical to reduce the computational cost of Algorithm \ref{Algorithm: Gradient Descent} by avoiding the full computation of EVD or even discarding them.

In relation to the original SDP problem in \eqref{SDP relaxation}, we can understand the PSD constraint in \eqref{SDP relaxation: psd constraint} is now penalized as the penalty term in $h_\gamma(\bu)$, i.e., $\frac{\gamma}{2}\|\Pi_+(\bC(\bu))\|_F^2$. Thus, one of the key insights we will use is that: i) if the PSD constraint is not satisfied, the penalty term equals to zero, simplifying the objective function as $h_\gamma(\bu)=\bu^T\bones$; in this case, the gradient is simply $\nabla_{\bu} h_\gamma(\bu)=\bones$, eliminating the computation of EVD, and ii) if the PSD constraint is satisfied, the penalty term becomes nonzero and it requires the computation of EVD to find out $\nabla_{\bu} h_\gamma(\bu)$. This fact leads to the following proposition showcasing the rule of updating $\bu_{i+1}$ for the enhanced GD.
\begin{proposition} \label{Proposition: Enhanced GD}
The enhanced GD includes two cases depending on the condition of the PSD constraint as
\begin{align}\label{two-phase GD}
\textstyle
\begin{cases}
  \text{Case A:} ~ \mbox{if the PSD constraint does not meet} \\
  \quad \quad \quad \quad \quad \quad \quad \quad \quad \quad \Rightarrow    \bu_{i+1} = \bu_{i} - t_i\bones \\
  \text{Case B:} ~ \mbox{if the PSD constraint meets} \\
  \quad \quad \quad \quad \quad \quad \quad \quad \quad \quad \Rightarrow    \bu_{i+1} = \bu_{i} - t_i\nabla_{\bu_i} h_\gamma(\bu_i)
\end{cases}. \nn
\end{align}
\end{proposition}
The key is to check if the PSD constraint in Proposition \ref{Proposition: Enhanced GD} is satisfied or not without the need of computing EVD. We propose a simple sufficient condition, based on the Weyl's inequality \cite{Horn1991}, as demonstrated in the proposed Algorithm \ref{Algorithm: Enhanced Gradient Descent}.

In Algorithm \ref{Algorithm: Enhanced Gradient Descent}, we focus on modifying Step 2 in Algorithm \ref{Algorithm: Gradient Descent} by using an initial $\bu_0$ with equal entries (for instance, $\bu_0=\bones$) and exploiting the fact that $\nabla h_\gamma(\bu_i)=\bones$ if $\bC(\bu_i)$ is not PSD (Case A in Proposition \ref{Proposition: Enhanced GD}) to reduce the computational cost of EVD. Step $4$ in Algorithm \ref{Algorithm: Enhanced Gradient Descent} is due to the fact that the $k$th eigenvalue of $\bA+\alpha\bI~(\alpha\in\R)$ is $\lambda_k(\bA)+\alpha$. One of the key insights we leverage is that the choice of the sequence of gradient directions, i.e., $\nabla_{\bu_i} h_\gamma(\bu_i)$, $i=0,1,\ldots$, does not alter the optimality of the dual problem in \eqref{convex transformation}. We approach the design of $\bu_0$ with the goal of eliminating the computation of EVD to the most extent. Moreover, in Step $11$ of Algorithm \ref{Algorithm: Enhanced Gradient Descent}, the condition $\lambda_{\text{max}}(-\bA) \!+\! \lambda_{\text{max}}(-\text{diag}(\bu_i))\leq 0  \Rightarrow  \lambda_{\text{max}}(\bC(\bu_i))\leq 0$ (Case A in Proposition \ref{Proposition: Enhanced GD}), holds because of the Weyl's inequality \cite{Horn1991}.
Note that we accelerate the original GD by reducing the computation of EVD from two different perspectives: one is from a better designed initial point $\bu_0$ and another one is taking into acount the charateristics of $\bC(\bu_i)$, i.e., $\bC(\bu_i)$ is PSD or not.
The EVD of $\!\bC(\bu_i)\!$ is required only when both the conditions ``\mbox{all} the elements of $\bu_i$ are the same" and ``$\lambda_{\text{max}}(-\bA) + \lambda_{\text{max}}($ $-\text{diag}(\bu_i))\leq 0$" are violated, as in \emph{Phase II-B}.
The effectiveness of the proposed enhanced GD will be validated by using numerical results in Section \ref{Sec.5}.

\begin{figure}[t]
\centering
\includegraphics[width=7.2cm, trim=10 05 10 05, clip]{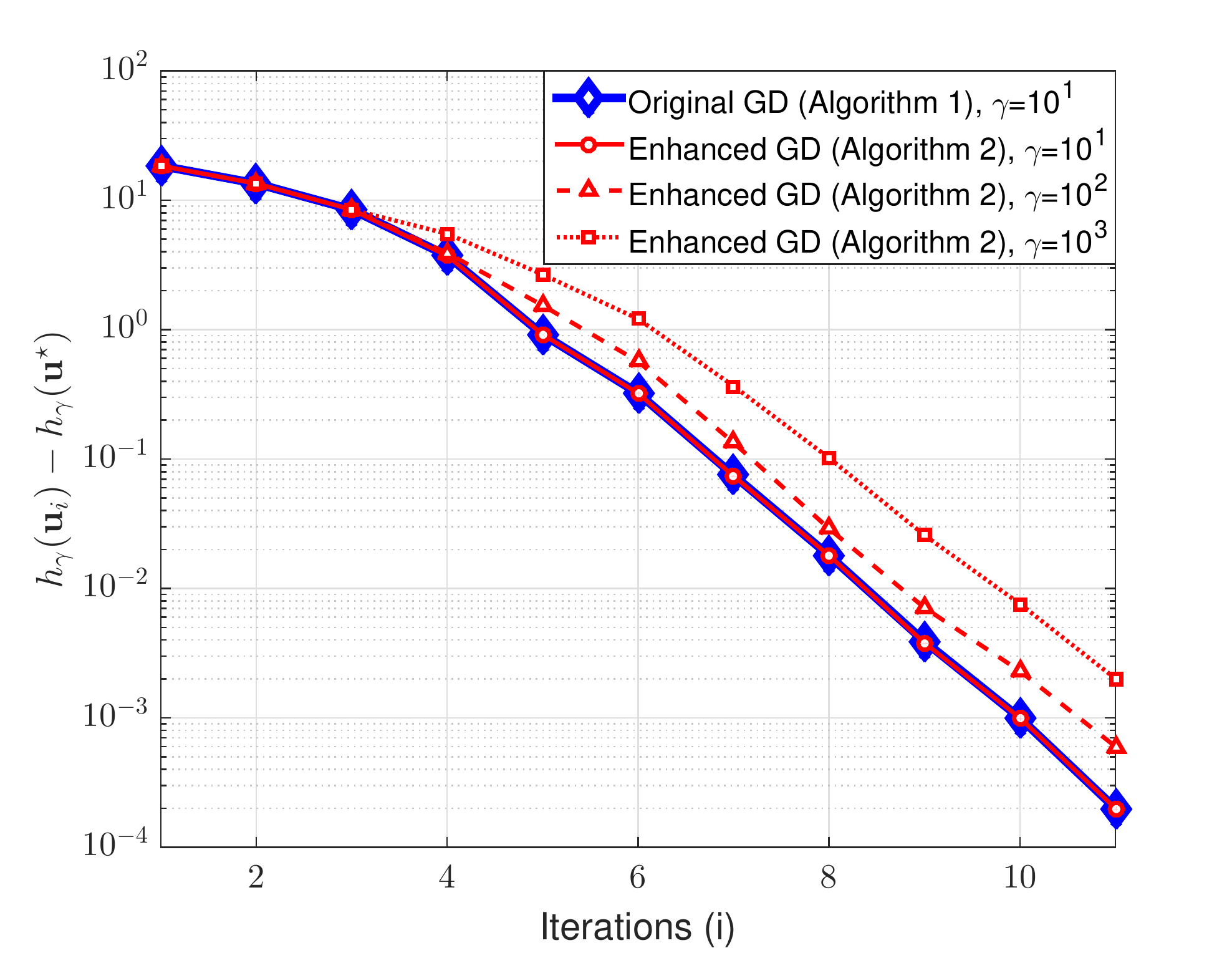}
\caption{Convergence rate comparison of GD in Algorithm \ref{Algorithm: Gradient Descent} and the enhanced GD with EVD elimination in Algorithm \ref{Algorithm: Enhanced Gradient Descent} when $D=4$.} \label{Convergencefig1}
\end{figure}

Notice that Step 2 in computing $\nabla_{\bu_i} h_\gamma(\bu_i)$ of Algorithm \ref{Algorithm: Gradient Descent} has been transformed into two different phases (each phase includes two sub-phases) in Algorithm \ref{Algorithm: Enhanced Gradient Descent}. Algorithm \ref{Algorithm: Enhanced Gradient Descent} executes the four sub-phases in order and irreversibly. To be specific, the algorithm first enters \emph{Phase I} at the initial iteration and ends up with \emph{Phase II}. Once the algorithm enters \emph{Phase II}, there is no way to return back to \emph{Phase I}. The duration of four sub-phases varies with the characteristics of $\bC(\bu_i)$, which depends on $D$ and the dataset. An example will be taken to illustrate the duration of phases in Algorithm \ref{Algorithm: Enhanced Gradient Descent} in Section \ref{Sec.5.1}. Algorithms \ref{Algorithm: Gradient Descent} and \ref{Algorithm: Enhanced Gradient Descent} are based on GD, and thus the enhanced GD does not alter the convergency of Algorithm \ref{Algorithm: Gradient Descent} \cite{Boyd04}.
\begin{proposition}[Convergence Rate of the Enhanced GD]
Let $\bu^\star$ be the optimal solution to the strongly convex problem in \eqref{convex transformation}. Then if we run Algorithm \ref{Algorithm: Enhanced Gradient Descent} for $i$ iterations, it will yield a solution $h_\gamma(\bu_{i})$ which satisfies
\beq
h_\gamma(\bu_{i}) - h_\gamma(\bu^\star) \leq \cO(c^i), ~ 0<c<1, ~ i=1,2,\ldots \nn
\eeq
Intuitively, this means that the enhanced GD is guaranteed to converge with the convergence rate $\cO(c^i)$.
\end{proposition}

\begin{remark}
Both Algorithms \ref{Algorithm: Gradient Descent} and \ref{Algorithm: Enhanced Gradient Descent}, which are based on the original GD \cite{Boyd04,Kim16}, result in the same update sequences $\{\bu_i\}$. This phenomenon is captured in Fig. \ref{Convergencefig1}, in which the optimality gap (i.e., $h_\gamma(\bu_i)-h_\gamma(\bu^\star)$) as a function of the iteration number $i$ is depicted for Algorithms \ref{Algorithm: Gradient Descent} and \ref{Algorithm: Enhanced Gradient Descent}. In terms of flops, however, Algorithm \ref{Algorithm: Enhanced Gradient Descent} is more efficient than Algorithm \ref{Algorithm: Gradient Descent}. This leads to a dramatic reduction in the running time of Algorithm \ref{Algorithm: Enhanced Gradient Descent} since we mainly move on the direction obtained without the computation of EVD.
Furthermore, the asymptotic error bound in Proposition \ref{Proposition: Enhanced GD} is unassociated with $\gamma$, in which the bound converges to zero as $i$ tends to infinity. This asymptote is captured by the slop of the error decrease as $\log(h_{\gamma}(\bu_i)-h_{\gamma}(\bu^\star))\leq \cO(\log(c)\cdot i)$, where $\log(c)<0$ defines the asymptotic slop and is independent of $\gamma$. We utilize simulation curves to show the effect of $\gamma$ on the convergence rate of the enhanced GD in Fig. \ref{Convergencefig1}. It can be observed that a larger $\gamma$ leads to a slower convergence (i.e., a larger shift of the red curves to the right). Nevertheless, $\gamma$ needs to be chosen by considering the tradeoff between the training performance of KM and the computational cost as illustrated in Section \ref{Sec.5.1}.
\end{remark}

\begin{algorithm}[t]
\caption{Randomization} \label{Algorithm: Randomization}
\begin{algorithmic}[1]
\Require
$\bA$, $\Pi_+(\bC(\bu^\star))=\bV_{+}\bLam_{+}\bV_{+}^T$, $D$, $\gamma$, and $I_{\text{rand}}$ (\emph{the number of randomizations}).
\Ensure
$\hat{\bpsi}$ (an approximate solution to the BQP in \eqref{subproblem 2: BQP}).
\State Obtain $\bL = \bV_+\sqrt{\gamma\bLam_+}$ and $\bL\bL^T=\bX^\star$.
\For{$\ell=1, 2, \ldots, I_{\text{rand}}$}
\State Generate an independent and identically distributed (i.i.d.) Gaussian random vector: $\bxi_\ell\sim\cN(\bzeros, \bI_{D+1})$.
\State Random sampling: $\tilde{\bxi}_\ell = \bL\bxi_\ell$.
\State Discretization: $\tilde{\bx}_\ell = \text{sign}(\tilde{\bxi}_\ell)$.
\EndFor
\State Determine $\ell^\star = \argmin_{\ell=1,\ldots,I_{\text{rand}}} \tilde{\bx}_\ell^T\bA\tilde{\bx}_\ell$.
\State Approximation: $\hat{\bx} = \tilde{\bx}_{\ell^\star}(1)\cdot \tilde{\bx}_{\ell^\star}(2:D+1)$ and $\hat{\bpsi}=(\hat{\bx} + \bones)/2$.
\end{algorithmic}
\end{algorithm}

\subsection{Randomization}
The solution to the dual problem in \eqref{convex transformation} (or equivalently \eqref{eq:dual}) produced by Algorithm \ref{Algorithm: Enhanced Gradient Descent}, is not yet a feasible solution to the BQP in \eqref{subproblem 2: BQP}. A randomization procedure \cite{Goemans1995} can be employed to extract a feasible binary solution to \eqref{subproblem 2: BQP} from the SDP solution $\bX^\star$ of \eqref{regularized SDP}. One typical design of the randomization procedure for BQP is to generate feasible points from the Gaussian random samples via rounding \cite{Luo10}. The Gaussian randomization procedure provides a tight approximation with probability $1-\exp(-\cO(D))$, asymptotically in $D$ \cite{Kisialiou05}. By leveraging the fact that the eigenvalues and corresponding eigenvectors of $\Pi_+(\bC(\bu))$ can be found by Steps 13 and 14 of Algorithm \ref{Algorithm: Enhanced Gradient Descent}, we have
\beq
\bX^\star = \gamma\Pi_+(\bC(\bu^\star)) = \gamma \bV_{+}\bLam_{+}\bV_{+}^T = \bL\bL^T, \nn
\eeq
where the first equality follows from \eqref{optimal solution of SDP relaxation-0} and \eqref{eq:useful equality}, $\Pi_+(\bC(\bu))$ $\triangleq \bV_{+}\bLam_{+}\bV_{+}^T$, and $\bL = \bV_+\sqrt{\gamma\bLam_+}$. A detailed randomization procedure is provided in Algorithm \ref{Algorithm: Randomization}.

In Step 8 of Algorithm \ref{Algorithm: Randomization}, the $D$-dimensional vector $\hat{\bx}$ is first recovered from a $(D+1)$-dimensional vector $\tilde{\bx}_{\ell^\star}$ by considering the structure of $\bX^\star$ in \eqref{reformulated BQP+}, and then used to approximate the BQP solution based on \eqref{reformulated BQP}. Also note that the randomization performance improves with $I_{\text{rand}}$. In practice, we only need to choose a sufficient but not excessive $I_{\text{rand}}$ (for instance, $50\leq I_{\text{rand}} \leq 100$) achieving a good approximation for the BQP solution. Moreover, its overall computational complexity is much smaller than the conventional randomization algorithms \cite{Goemans1995,Luo10,Ghauch18} because our proposed Algorithm \ref{Algorithm: Randomization} does not require the computation of the Cholesky factorization.

\begin{algorithm}[t]
\caption{Dual Optimization for KM learning with Enhanced GD} \label{Algorithm: Overall KM learning}
\begin{algorithmic}[1]
\Require
$\cU_{\cK}$, $\cI_{\cK}$, $\cK$, $\{p_{u,i}\}_{(u,i)\in\cK}$, and $I_{\text{BCD}}$. Initialize $\{\bthe_u^{(1)}\in \cP\}_{u\in\cU_{\cK}}$.
\Ensure
$\{\bthe_u^\star\}_{u\in\cU_{\cK}}$, $\{\bpsi_i^\star\}_{i\in\cI_{\cK}}$.
\For{$\tau = 1, 2, \ldots, I_{\text{BCD}}$}
\State Update $\{\bpsi_i^{(\tau)}\}_{i\in\cI_{\cK}}$: %the binary indicator vectors
       \For{$i\in\cI_{\cK}$}
       \State Obtain $\bu_i^\star$ from Algorithm \ref{Algorithm: Enhanced Gradient Descent}.
       \State Recover $\bpsi_i^{(\tau)}$ from Algorithm \ref{Algorithm: Randomization}.
       \EndFor
\State Update $\{\!\bthe_u^{(\tau)}\!\}_{\!u\in\cU_{\cK}}$: %the probability mass function vectors
       \For{$u\in\cU_{\cK}$}
       \State Obtain $\bthe_u^{(\tau)}$ from the FW algorithm \cite{Jaggi13}.
       \EndFor
\EndFor \\
\Return $\{\bthe_u^\star = \bthe_u^{(I_{\text{BCD}})}\}_{u\in\cU_{\cK}}$ and $\{\bpsi_i^\star = \bpsi_i^{(I_{\text{BCD}})}\}_{i\in\cI_{\cK}}$.
\end{algorithmic}
\end{algorithm}

\subsection{Overall KM Learning Algorithm}
Incorporating Algorithm \ref{Algorithm: Enhanced Gradient Descent} and Algorithm \ref{Algorithm: Randomization}, the overall KM learning framework is described in Algorithm \ref{Algorithm: Overall KM learning}.

Note that the index of BCD iterations $\tau$ that has been omitted is recovered here and $I_{\text{BCD}}$ denotes the total number of BCD iterations for KM learning. In Algorithm \ref{Algorithm: Overall KM learning}, the BCD method is adopted to refine $\{\bpsi_i^{(\tau)}\}_{i\in\cI_{\cK}}$ and $\{\bthe_u^{(\tau)}\}_{u\in\cU_{\cK}}$ until it converges to a stationary point of \eqref{KM problem formulation}. In fact, the proof of convergence (to stationary solution) for Algorithm \ref{Algorithm: Overall KM learning} is exactly the same as that of Algorithm 1 in \cite{Ghauch18}. In practice, we can use $I_{\text{BCD}}$ to control the termination of Algorithm \ref{Algorithm: Overall KM learning}.

\section{Approximate EVD and Error Analysis}
In this section,  several techniques are discussed to further accelerate Algorithm \ref{Algorithm: Enhanced Gradient Descent}.

\subsection{Initial Step Size}
\begin{figure}[t]
\centering
\includegraphics[width=7.2cm]{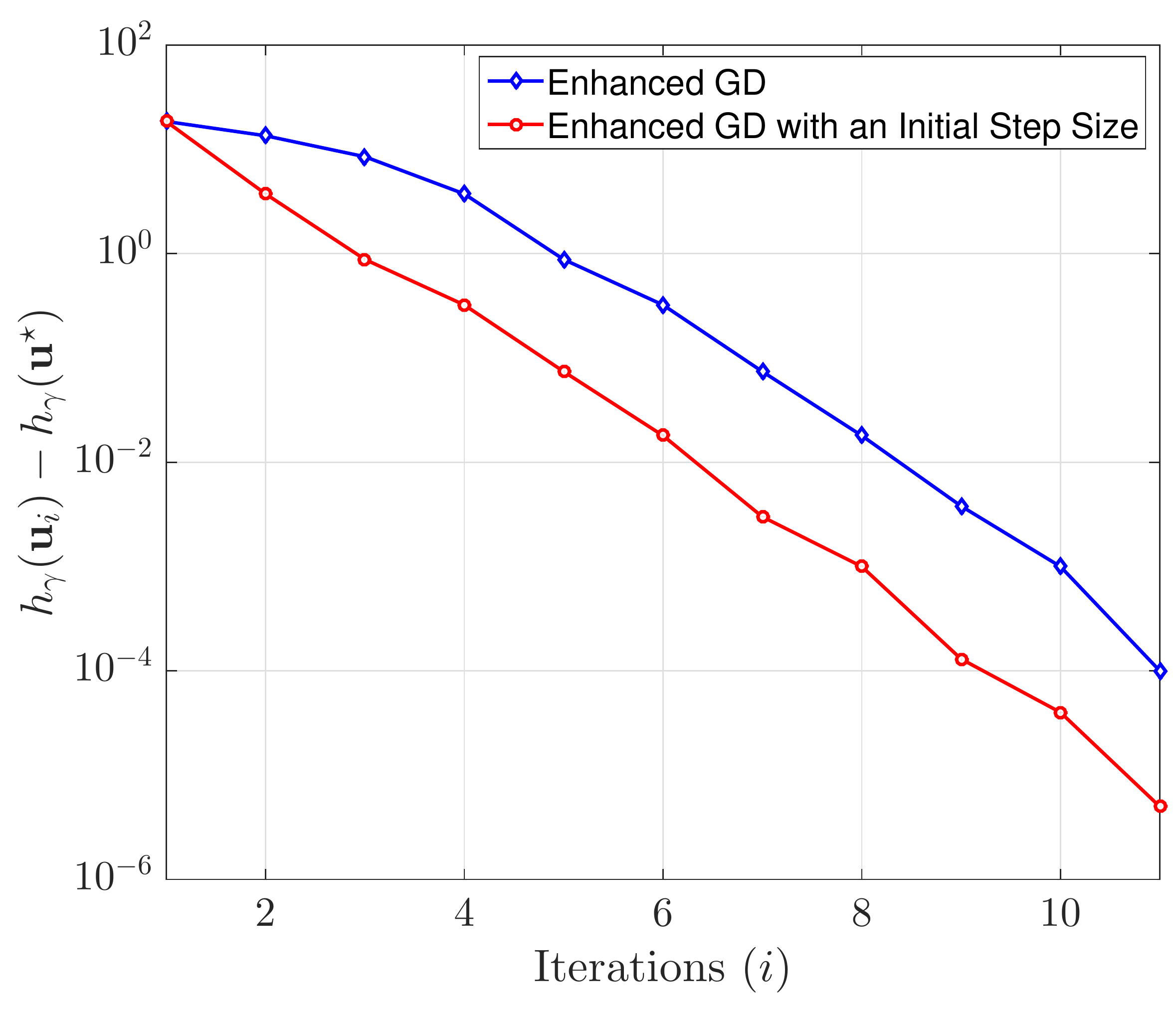}
\caption{Convergence rate comparison of the enhanced GD with EVD elimination in Algorithm \ref{Algorithm: Enhanced Gradient Descent} and that with an initial step size when $D=4$.} \label{Convergencefig2}
\end{figure}

A good initial step size $t_0$ is crucial for the convergence speed of the enhanced GD. In Phase I-A of Algorithm \ref{Algorithm: Enhanced Gradient Descent}, we have
\beq
\blam(\bC(\bu_{i+1})) = \blam(-\bA) - \bu_{i+1} = \blam(-\bA) - \bu_i + t_i\bones. \nn
\eeq
If $\lambda_{\text{max}}(\bC(\bu_{i+1}))>0$, the following holds $t_i > u_i - \lambda_{\text{max}}(-\bA)$ where $u_i \triangleq u_i(1) = \cdots = u_i(D+1)$. Therefore, in the first iteration of Phase I-A, we can set an appropriate step size $t_0 > u_0 - \lambda_{\text{max}}(-\bA)$ so that $\bC(\bu_1) = -\bA - \text{diag}(\bu_1)$ has at least one positive eigenvalue, where $\bu_1 = \bu_0 - t_0\bones$. With the above modification of Algorithm \ref{Algorithm: Enhanced Gradient Descent}, we can reduce the execution time spent in Phase I-A, and thus, the total number of iterations required by the enhanced GD can be reduced as shown in Fig. \ref{Convergencefig2}. Moreover, the choice of $\bu_0$ does not affect the overall performance in terms of the computational cost.

\subsection{Approximate EVD}
Compared to the original GD in Algorithm \ref{Algorithm: Gradient Descent}, the enhanced GD in Algorithm \ref{Algorithm: Enhanced Gradient Descent} has reduced the costly EVD substantially. Nevertheless, the EVD is still necessary in Algorithm \ref{Algorithm: Enhanced Gradient Descent} when the algorithm enters into \emph{Phase II-B}. In order to further accelerate the algorithm, we employ and modify the Lanczos method to numerically compute the approximate EVD of $\bC(\bu_i)$ in Algorithm \ref{Algorithm: Enhanced Gradient Descent}.

The Lanczos algorithm \cite{Lanczos1950} is a special case of the Arnoldi method \cite{Arnoldi1951} when the matrix is symmetric. In principle, it is based on an orthogonal projection of $\bC(\bu_i)$ onto the Krylov subspace $\cK_m\triangleq \text{span}\{\bp, \bC(\bu_i)\bp, \cdots, {\bC(\bu_i)}^{m-1}\bp\}$ where $m$ denotes the dimension of Krylov subspace. An algorithmic description of a modified Lanczos method is presented in Algorithm \ref{Algorithm: Lanczos method}.

\begin{algorithm}[t]
\caption{Modified Lanczos Algorithm} \label{Algorithm: Lanczos method}
\begin{algorithmic}[1]
\Require
$\bC(\bu_i)$, $D$, and $\delta$ (\emph{threshold value}). Choose an initial unit-norm vector $\bp_1\in\R^{D+1}$. Set $\beta_1=0$, $\bp_0=\bzeros$, and $\bH_{m}=\bzeros_{(D+1)\times (D+1)}$ ($\bzeros_{(D+1)\times (D+1)}$ denotes the all-zero matrix of dimension $(D+1)\times (D+1)$).
\Ensure
$\bP_m=[\bp_1,\bp_2,\cdots, \bp_m]$ and $\bH_m$. %=[h_{i,j}]$~($h_{i,j}$ denotes the $(i,j)$th entry of $\bH_m$).
\For{$j=1, 2, \ldots, D+1$}
 \State $\bw_j = \bC(\bu_i)\bp_j - \beta_j\bp_{j-1}$.
 \State $\alpha_j = \langle \bw_j, \bp_j \rangle$ and $H_m(j,j) = \alpha_j$ ($\alpha_j$ forms the main diagonal of $\bH_m$).
 \State $\bw_j = \bw_j - \alpha_j\bp_j$ and $\beta_{j+1} = \|\bw_j\|_2$.
 \If {$\beta_{j+1}\leq \delta$} terminate and  \label{modified Lanczos:stopping condition}
 \Return
 \State $m = j$ and $\bH_m = \bH_m(1:m, 1:m)$.
 \Else
 \State $H_m(j,j+1)=H_m(j+1,j)=\beta_{j+1}$ ($\beta_{j+1}$ forms the super- and sub-diagonal of $\bH_m$).
 \State $\bp_{j+1} = \bw_j/\beta_{j+1}$.
 \EndIf
\EndFor
\end{algorithmic}
\end{algorithm}

Different from the Arnoldi method, the matrix $\bH_m\in\R^{m\times m}$ constructed by Algorithm \ref{Algorithm: Lanczos method} is tridiagonal and symmetric, i.e., the entries of $\bH_m$ in Algorithm \ref{Algorithm: Lanczos method} satisfy that $H_m(i,j) = 0$, $1 \leq i < j-1$, and $H_m(j,j+1) = H_m(j+1,j)$, $j = 1,2,\ldots,m$.
Also, Algorithm \ref{Algorithm: Lanczos method} iteratively builds an orthonormal basis, i.e., $\bP_m\in\R^{(D+1)\times m}$,  for $\cK_m$ such that $\bP_m^T\bC(\bu_i)\bP_m=\bH_m$ and $\bP_m^T\bP_m=\bI_m$, where $m\leq D+1$. Let $(\vartheta_i, \bq_i)$, $i=1,\ldots,m$, be the eigenpairs of $\bH_m$. Then, the eigenvalues/eigenvectors of $\bC(\bu_i)$ can be approximated by the Ritz pairs $(\vartheta_i, \bP_m\bq_i)$, i.e.,
\beq \label{eq: Rayleigh-Ritz method}
\hat{\lambda}_i = \vartheta_i, \ \hat{\bv}_i = \bP_m\bq_i, \ i=1,\ldots,m.
\eeq

With the increase of the dimension of Krylov subspace $m$, the approximation performance improves at the price of additional computations. Thus, in practice, we adopt the value of $m$ balancing the tradeoff between the accuracy of approximation and the computational complexity.

\subsection{Analysis of Approximation Error and Thresholding Scheme} \label{Sec.4.3}
In this subsection, we analyze the approximation error of the approximate EVD and propose a thresholding scheme for selecting an appropriate $m$ in Algorithm \ref{Algorithm: Lanczos method}. The main results are provided in the following lemmas.
\begin{lemma} \label{Lemma:Approximation error}
Let $(\vartheta_i, \bq_i)$ be any eigenpair of $\bH_m$ and $(\hat{\lambda}_i = \vartheta_i, \hat{\bv}_i = \bP_m\bq_i)$ in \eqref{eq: Rayleigh-Ritz method} is an approximated eigenpair (Ritz pair) of $\bC(\bu_i)$ in Algorithm \ref{Algorithm: Lanczos method}. Then the following holds:

i) The residual error $r_e(\bC(\bu_i)\hat{\bv}_i, \hat{\lambda}_i\hat{\bv}_i)\triangleq \|\bC(\bu_i)\hat{\bv}_i - \hat{\lambda}_i\hat{\bv}_i\|_2$ is upper bounded by
\beq \label{upper bound of approximation error}
r_e(\bC(\bu_i)\hat{\bv}_i, \hat{\lambda}_i\hat{\bv}_i) \leq \beta_{m+1}.
\eeq

ii) The maximum approximation error of eigenvalues of $\bC(\bu_i)$ is bounded by
   \beq \label{eigenvalue difference}
   \max_i |\lambda_i - \hat{\lambda}_i| \leq \beta_{m+1}, \ i\in\{1,\ldots,m\},
   \eeq
where $\lambda_i$ is the associated true eigenvalue of $\bC(\bu_i)$.

iii) The minimum approximation error of eigenvalues of $\bC(\bu_i)$ is bounded by
   \beq \label{minimum approximation error}
   \min_i |\lambda_i - \hat{\lambda}_i|\leq \beta_{m+1}|q_i(m)|, \ i\in\{1,\ldots,m\}.
   \eeq
\end{lemma}
\quad\textit{Proof:} See Appendix \ref{Proof of Lemma 2}.

Lemma \ref{Lemma:Approximation error} indicates that the error bounds of the approximate eigenvalues of $\bC(\bu_i)$ by using the approximate EVD in Algorithm \ref{Algorithm: Lanczos method} largely depends on $\beta_{m+1}$. Indeed, the upper bounds in \eqref{eigenvalue difference} and \eqref{minimum approximation error} are quite tight as will be seen in Section \ref{Sec.5}. Inspired by Lemma \ref{Lemma:Approximation error}, finding an upper bound of $\beta_{m+1}$ that only depends on the trace of $\bC(\bu_i)$ is of interest.

\begin{lemma} \label{Lemma:UB on beta}
$\beta_{m+1}$ in Algorithm \ref{Algorithm: Lanczos method} is upper bounded by
\beq \label{upper bound on beta}
\beta_{m+1} \leq 2m \big((\sigma_{\text{max,UB}}-\sigma_{\text{max,LB}}) + \hat{\sigma}_{\text{max,Minkowski}}\big),
\eeq
where $\sigma_{\text{max,UB}} \!=\! \frac{\text{trace}(\bC(\bu_i))}{D+1} \!+\! ((\frac{\text{trace}(\bC(\bu_i)^2)}{D+1}\!-\!\frac{\text{trace}^2(\bC(\bu_i))}{(D+1)^2})\cdot D)^{\frac{1}{2}}$ and $\sigma_{\text{max,LB}} \!=\! \frac{\text{trace}(\bC(\bu_i))}{D+1} + ((\frac{\text{trace}(\bC(\bu_i)^2)}{D+1}-\frac{\text{trace}^2(\bC(\bu_i))}{(D+1)^2})/ D)^{\frac{1}{2}}$ are the upper and lower bounds on the largest singular value of $\bC(\bu_i)$, respectively\footnote{For the symmetric matrix $\bC(\bu_i)\in\S^{(D+1)\times(D+1)}$, its largest singular value $\sigma_{\text{max}}(\bC(\bu_i))$ is the same as the absolute value of its eigenvalue with the largest modulus.}, and $\hat{\sigma}_{\text{max,Minkowski}}\triangleq \frac{1}{D+1}$ $\sum_{\ell=1}^{D+1}\sum_{j=1}^{D+1}|C(\ell,j)|$ is a normalized Minkowski $\ell_1$-norm of $\bC(\bu_i)$ ($C(\ell,j)$ denotes the $(\ell,j)$th entry of $\bC(\bu_i)$ for simplicity).
\end{lemma}
\quad\textit{Proof:} See Appendix \ref{Proof of Lemma 3}.

\begin{table}[t]
	\centering
	\caption{Time Consumption (in Seconds) Comparison of Solving the BQP under Different $\gamma$ for (D1) when $D=16$} \label{Table:gamma}
    \resizebox{\linewidth}{!}{
	\begin{tabular}{|c|c|c|c|}
		\hline
		\diagbox{Algorithm}{$\gamma$} & $10^1$ & $10^2$ & $10^3$  \\
		\hline
		Original GD & $6.51\times 10^{-1}$ & $1.24$ & $2.79$ \\
		Enhanced GD & $6.79\times 10^{-2}$ & $1.53\times 10^{-1}$ & $2.62\times 10^{-1}$ \\
		\hline
	\end{tabular}}
\end{table}

\begin{figure}[t]
\centering
\includegraphics[width=7.2cm, trim=15 05 40 25, clip]{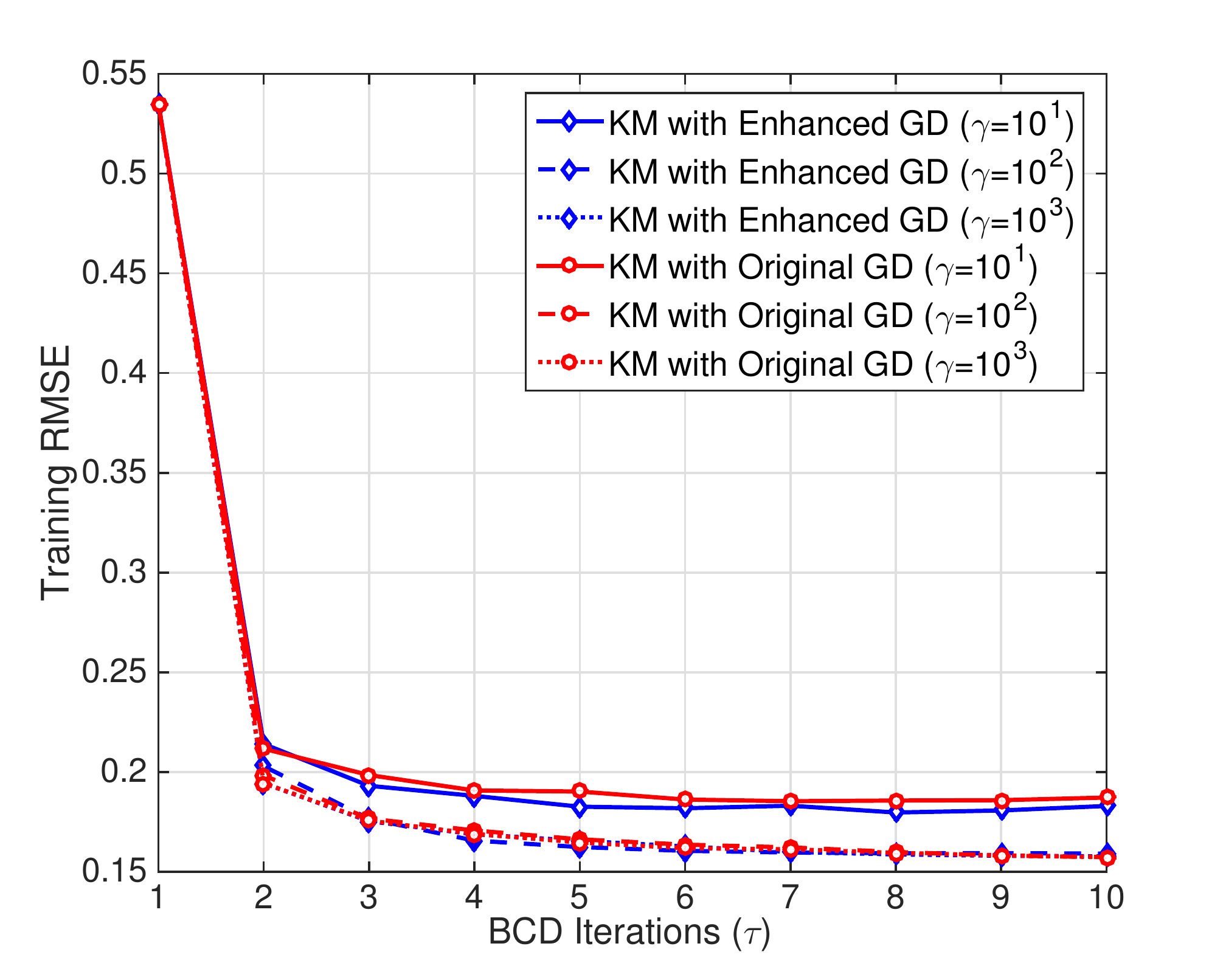}
\caption{The effect of $\gamma$ in \eqref{regularized SDP} on the KM training performance when $D=16$.} \label{Fig3:gamma}
\end{figure}

Lemma \ref{Lemma:UB on beta} gives us an upper bound of $\beta_{m+1}$ that does not require a computation of EVD and can be readily employed as a stopping condition in Step \ref{modified Lanczos:stopping condition} of Algorithm \ref{Algorithm: Lanczos method}. In particular, it proposes to use the normalized Minkowski $\ell_1$-norm $ \hat{\sigma}_{\text{max,Minkowski}}$. Notice that we introduced $\hat{\sigma}_{\text{max,Minkowski}}$ in Appendix C (Proof of Lemma \ref{Lemma:UB on beta}) to further upper bound $\sigma_{\text{max}}(\bC(\bu_i))$ in \eqref{upper bound based on sigma}, which gives a good approximation of $\sigma_{\text{max}}(\bC(\bu_i))$. It is important to note that the upper bound in \eqref{upper bound on beta} depends only on the traces and the absolute value of entries of $\bC(\bu_i)$, whose computational cost is extremely low compared to that of EVD. Moreover, in Appendix C, a useful property of $\hat{\sigma}_{\text{max,Minkowski}}$ is leveraged, namely, $\sigma_{\text{max,LB}}\leq \hat{\sigma}_{\text{max,Minkowski}} \leq \sigma_{\text{max,UB}}$ \cite{Park17}.

Lemma \ref{Lemma:UB on beta} motivates us to adjust the number of iterations $m$ of Algorithm \ref{Algorithm: Lanczos method} by proposing a low-complexity yet reasonable threshold, which exploits the structure of the upper bound on $\beta_{m+1}$ in \eqref{upper bound on beta}.
Therefore, we propose to use a threshold value provided as
\beq \label{eq:threshold value}
\delta = \frac{1}{aD\ln D}((\sigma_{\text{max,UB}}-\sigma_{\text{max,LB}}) + \hat{\sigma}_{\text{max,Minkowski}}),
\eeq
where $a>0$ is a controlled parameter. Unlike the prior works which choose $m$ in a greedy manner, this thresholding scheme determines $m$ by controlling the approximation error below $\delta$, leading to a balance between the accuracy of approximation and the computational complexity. This will be further investigated in Section \ref{Sec.5}.

\section{Numerical Results} \label{Sec.5}
We now present the simulation results demonstrating the superiority of the proposed methods compared with the conventional KM (e.g., the KM with SDRwR \cite{Ghauch18} and DMO \cite{Duan20}) and existing data representation techniques (e.g., NNM \cite{Stark16}, MF \cite{Koren09},  and SVD++ \cite{Koren08}) in terms of the computational cost, training and prediction performance, and interpretability. Three datasets for experiments are mainly considered, including (D1) an artificially generated toy dataset (for training only): $\cK=\{(u,i)|u\in \{1,\ldots,20\}, i\in \{1,\ldots,40\}\}$ $(U=20, I=40)$ and $\{p_{u,i}\}_{(u,i)\in\cK}$ are independent and uniformly distributed on the unit interval $[0,1]$, (D2) the MovieLens 100K dataset\footnote{https://grouplens.org/datasets/movielens/100k/} (ML100K) with $U=943$ users and $I=1682$ movie items, and (D3) the MovieLens 1 million dataset (ML1M) with $U=6040$ users and $I=3900$ movie items. For latter two MovieLens datasets, we divide each one of both into $80\%$ for training and the remaining $20\%$ for testing. The empirical probabilities of the training set, i.e., $\{p_{u,i}\}$, are obtained by $p_{u,i} = r_{u,i}/r_{\text{max}}$, $(u,i)\in\cK$, as in \eqref{eq:empirical probability}.

\begin{table}[t]
	\centering
	\caption{The duration of four sub-phases in Algorithm 2 based on (D1) when $D=8$} \label{Table:duration}
    \resizebox{0.7\linewidth}{!}{
	\begin{tabular}{|c|c|c|c|c|}
		\hline
		Phase & I-A & I-B & II-A & II-B  \\
		\hline
		Duration $(\%)$ & $37$ & $9$ & $8$ & $46$ \\
		\hline
	\end{tabular}}
\end{table}

\begin{table}[t]
 \centering
 \caption{Time Consumption (in Seconds) Comparison of the KM Learning $(D=8)$} \label{Table:CC0}
 \resizebox{\linewidth}{!}{
 \begin{tabular}{|c|c|c|c|}
  \hline
  Dataset & \diagbox{Algorithm}{Subproblem} & 1. LCQP & 2. BQP  \\
  \hline
  \multirowcell{4}{(D1)} & SDRwR & $1.51\times 10^{-1}$ & $7.36$ \\
                         & DMO & $1.43\times 10^{-1}$ & $1.85$ \\
                         & Original GD & $1.41\times 10^{-1}$ & $1.12\times 10^{-1}$ \\
                         & Enhanced GD & $1.37\times 10^{-1}$ & $8.60\times 10^{-2}$ \\
  \hline
  \multirowcell{4}{(D2)} & SDRwR & $7.05$ & $3.08\times 10^{+2}$ \\
                         & DMO & $7.11$ & $7.80\times 10^{+1}$ \\
                         & Original GD & $7.11$ & $2.24\times 10^{+1}$ \\
                         & Enhanced GD & $7.10$ & $2.19\times 10^{+1}$ \\
  \hline
\end{tabular}}
\end{table}

\begin{figure}[t]
\centering
\subfigure[]{
\includegraphics[width=7.2cm, trim=10 05 40 25, clip]{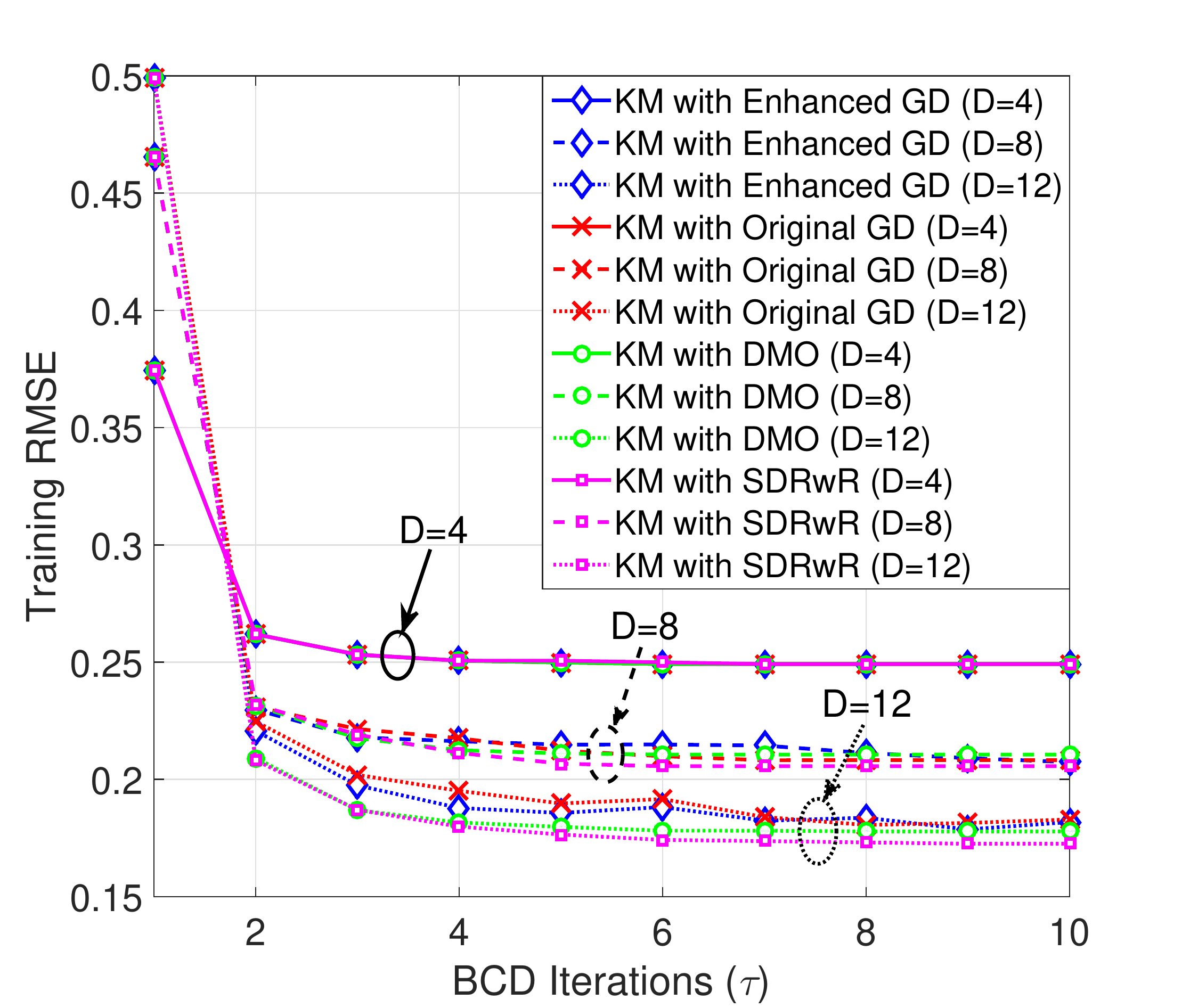} \label{Trainfig1-a}
}
\quad
\subfigure[]{
\includegraphics[width=7.2cm, trim=10 05 40 25, clip]{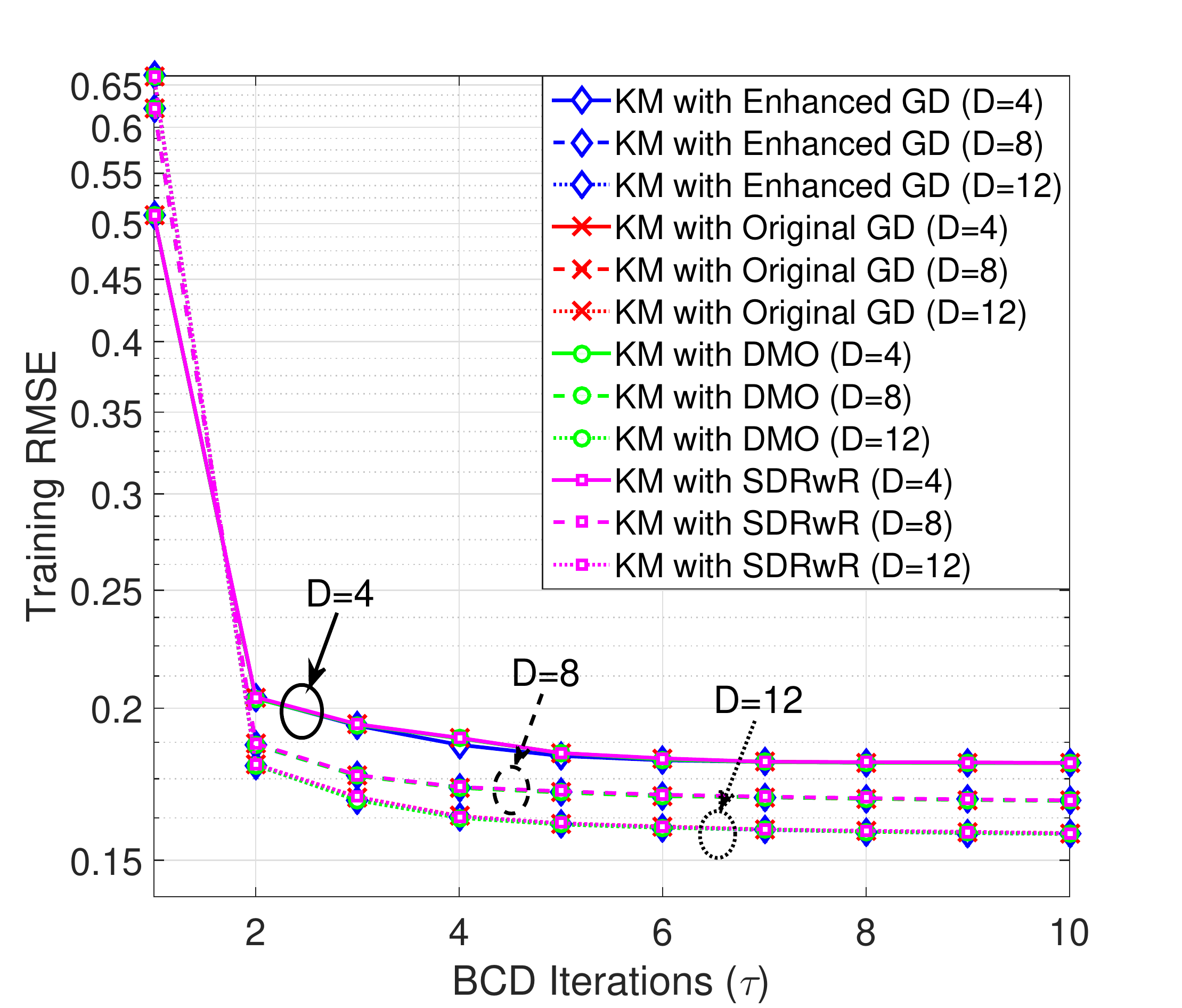} \label{Trainfig1-b}
}
\caption{Training RMSE vs. BCD iterations ($\tau$ in Algorithm \ref{Algorithm: Overall KM learning}): (a) the artificial dataset (D1), (b) ML100K (D2).} \label{Trainfig1}
\end{figure}

\subsection{Computational Cost and Training Performance} \label{Sec.5.1}
We evaluate the computational cost and training performance of the proposed KM learning with the enhanced GD (i.e., Algorithm \ref{Algorithm: Overall KM learning}). Throughout the paper, the computational cost is calculated by averaging the total running time in seconds (measured by ``cputime'' in MATLAB running on a PC with an Intel Core i7-7700 3.6 GHz CPU and 16 GB RAM) over the number of BCD iterations. We adopt the training root-mean-square error (RMSE), which is defined as $E_{\text{train}}\triangleq\sqrt{\frac{1}{|\cK|}\sum_{(u,i)\in\cK}|p_{u,i}-{\bthe^\star_u}^T\bpsi^\star_i|^2}$, as a metric.

We first investigate the effect of the regularization parameter $\gamma$ in \eqref{regularized SDP} on the KM learning performance. In Fig. \ref{Fig3:gamma}, the training RMSE is evaluated under different parameter settings of $\gamma$ for the two proposed GD-based methods based on the artificial dataset (D1). It can be seen from Fig. \ref{Fig3:gamma} that the larger the $\gamma$ value, the better the training performance of KM is, but this is achieved with an increased computational cost as shown in Table \ref{Table:gamma}. Moreover, it is indistinguishable in terms of the training error when $\gamma$ is increased from $10^2$ to $10^3$. It is thus a tradeoff between accuracy and complexity in the choice of $\gamma$. We choose $\gamma=100$ and fix it for subsequent numerical experiments.

In particular, we take the case when $D=8$ and (D1) is considered as an example to show the duration of each phase in the enhanced GD (Algorithm 2). In Table \ref{Table:duration}, the duration of each phase is measured by using the ratio between the number of iterations spent by the phase and the total number of iterations required by Algorithm 2 and averaged by taking $10^4$ realizations of (D1). Observed from Table \ref{Table:duration}, $54\%$ of iterations (including Phase I-A, I-B, and II-A) of Algorithm 2 do not require the computation of EVD, which results in a significant reduction of the computational cost compared to the original GD as will be shown in Table \ref{Table:CC1} (i.e., one or two orders of magnitude improvement in terms of the time overhead).

\begin{figure}[t]
\centering
\includegraphics[width=7.2cm, trim=05 00 20 20, clip]{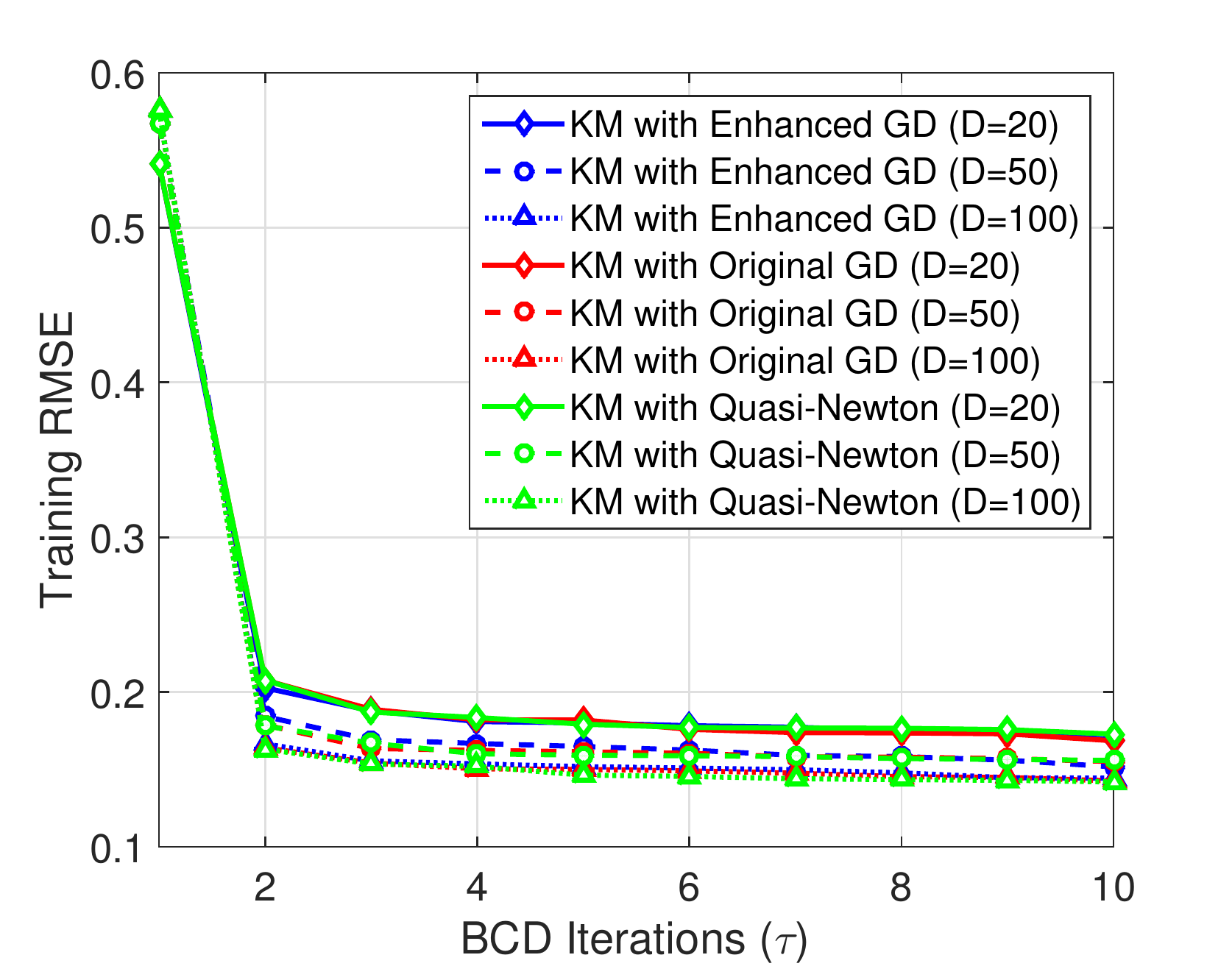}
\caption{KM training performance comparison between the quasi-Newton method and the proposed first-order methods based on (D1).} \label{Quasi-Newton}
\end{figure}

Table \ref{Table:CC0} demonstrates the computational complexity of the overall KM learning (LCQP + BQP) on the datasets (D1) and (D2), respectively, when $D=8$. In particular, the FW algorithm is fixed for solving the LCQP while different algorithms are applied to solving the BQP. It reveals that the computational cost of solving the LCQP in \eqref{subproblem 1: LCQP} via the FW algorithm is negligible compared to that of the BQP in \eqref{subproblem 2: BQP}, especially, via DMO or SDRwR. It can be seen that enhanced GD results in improved time complexity while it is clear that SDRwR and DMO are not scalable even for (D2), ML100K. The time complexity for solving the BQP with the varying $D$ can be found in Table \ref{Table:CC1}. Seen from Table \ref{Table:CC1}, the DMO shows benefits when $D$ is small, but its computational cost blows up as $D$ increases since the DMO is based on the branch-and-bound, which is very close to the exhaustive search in the worst case. For our proposed methods, the improvement on the computational cost of the enhanced GD compared to the original GD is marginal when $D$ is small. However, as $D$ grows, the benefit of the enhanced GD becomes significant. Fig. \ref{Trainfig1} displays the training RMSE comparison, which demonstrates that the proposed enhanced GD achieves similar, good training performance to the other approaches while reducing the computational complexity by several orders of magnitude as shown in Table \ref{Table:CC1}.
Furthermore, we compare the performance between the quasi-Newton method \cite{Shanno1970} and our proposed first-order methods based on (D1). It can be seen from Fig. \ref{Quasi-Newton} that the KM with quasi-Newton method achieves similar training performance as the gradient-based methods, while it consumes more computation time even compared with the original GD as shown in Table \ref{Table:CC1}. This is due to the fact that, in addition to calculating the gradient $\nabla_{\bu_i} h_\gamma(\bu_i)$ as in Algorithm \ref{Algorithm: Gradient Descent}, the quasi-Newton methods need to compute the approximated inverse of the Hessian matrix $\bH\approx(\nabla^2_{\bu_i} h_\gamma(\bu_i))^{-1}$ to obtain the descent direction ($\Delta\bu_i=-\bH\nabla_{\bu_i} h_\gamma(\bu_i)$).

\begin{figure}[t]
\centering
\includegraphics[width=6.8cm, trim=05 05 30 30, clip]{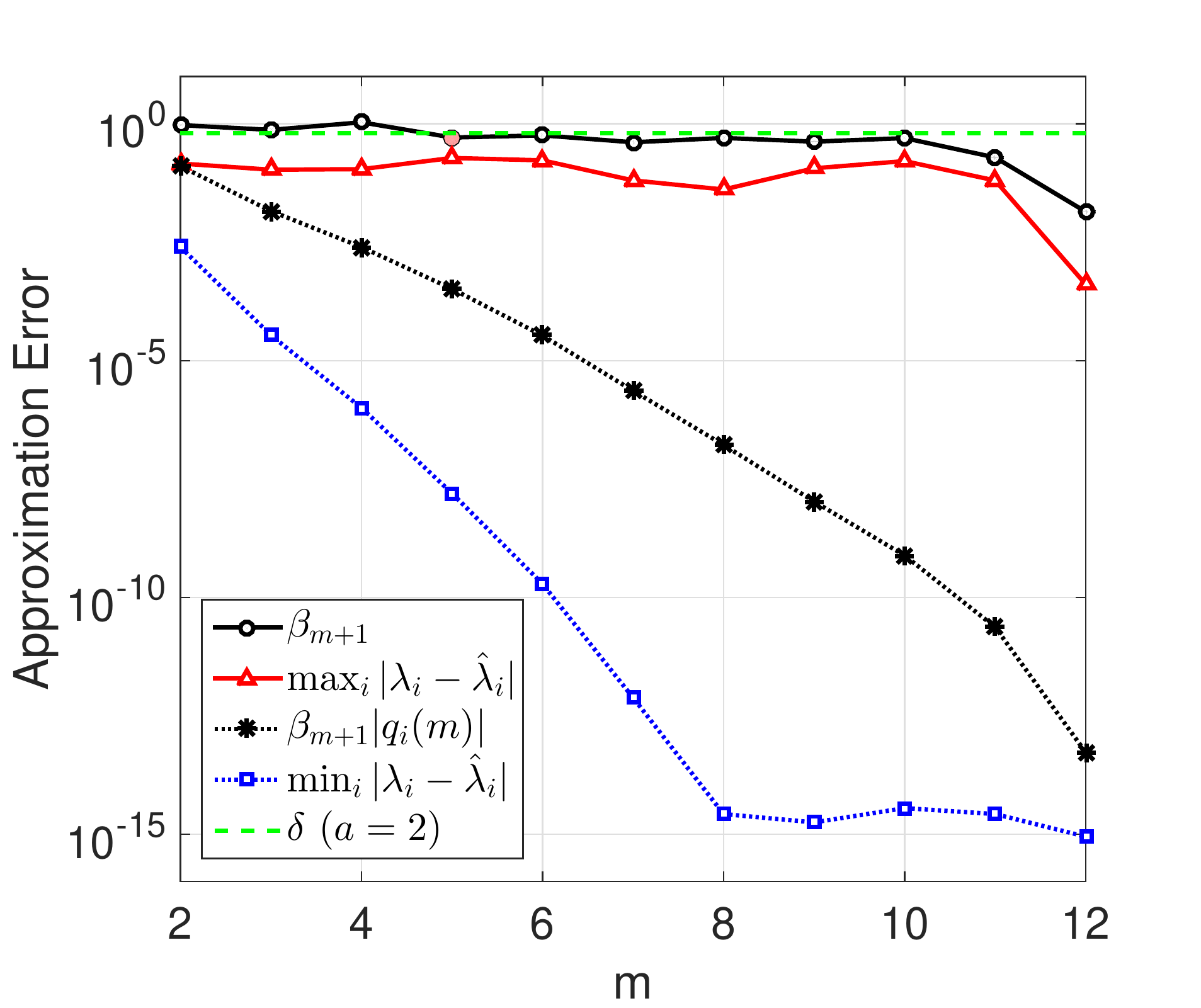}
\caption{Approximation error and upper bounds of the approximate EVD with thresholding when $D=12$.} \label{Approximation Error}
\end{figure}

\begin{figure}[t]
\centering
\subfigure[]{
\includegraphics[width=7.0cm, trim=10 0 40 20, clip]{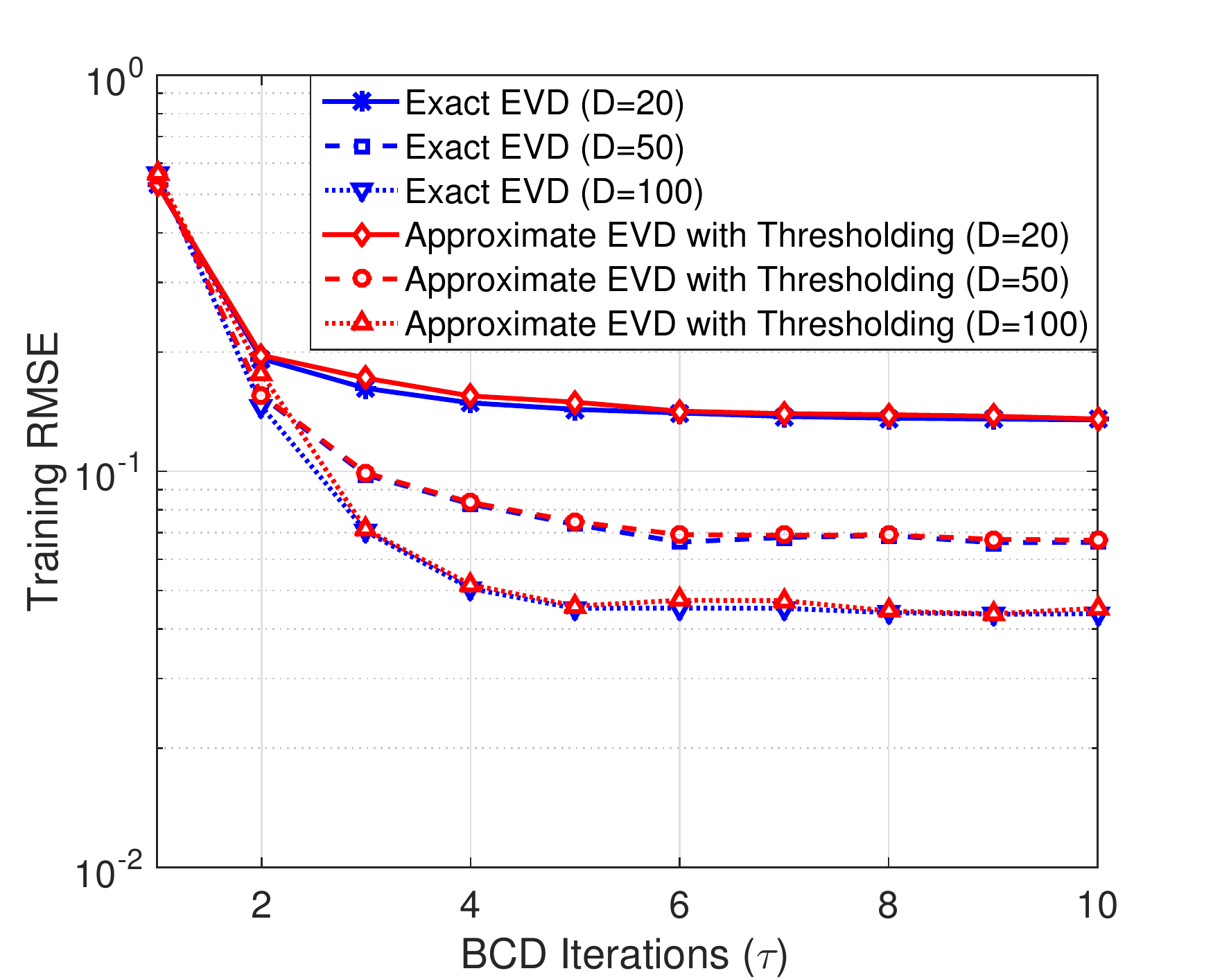}
}
\subfigure[]{
\includegraphics[width=7.0cm, trim=10 0 40 20, clip]{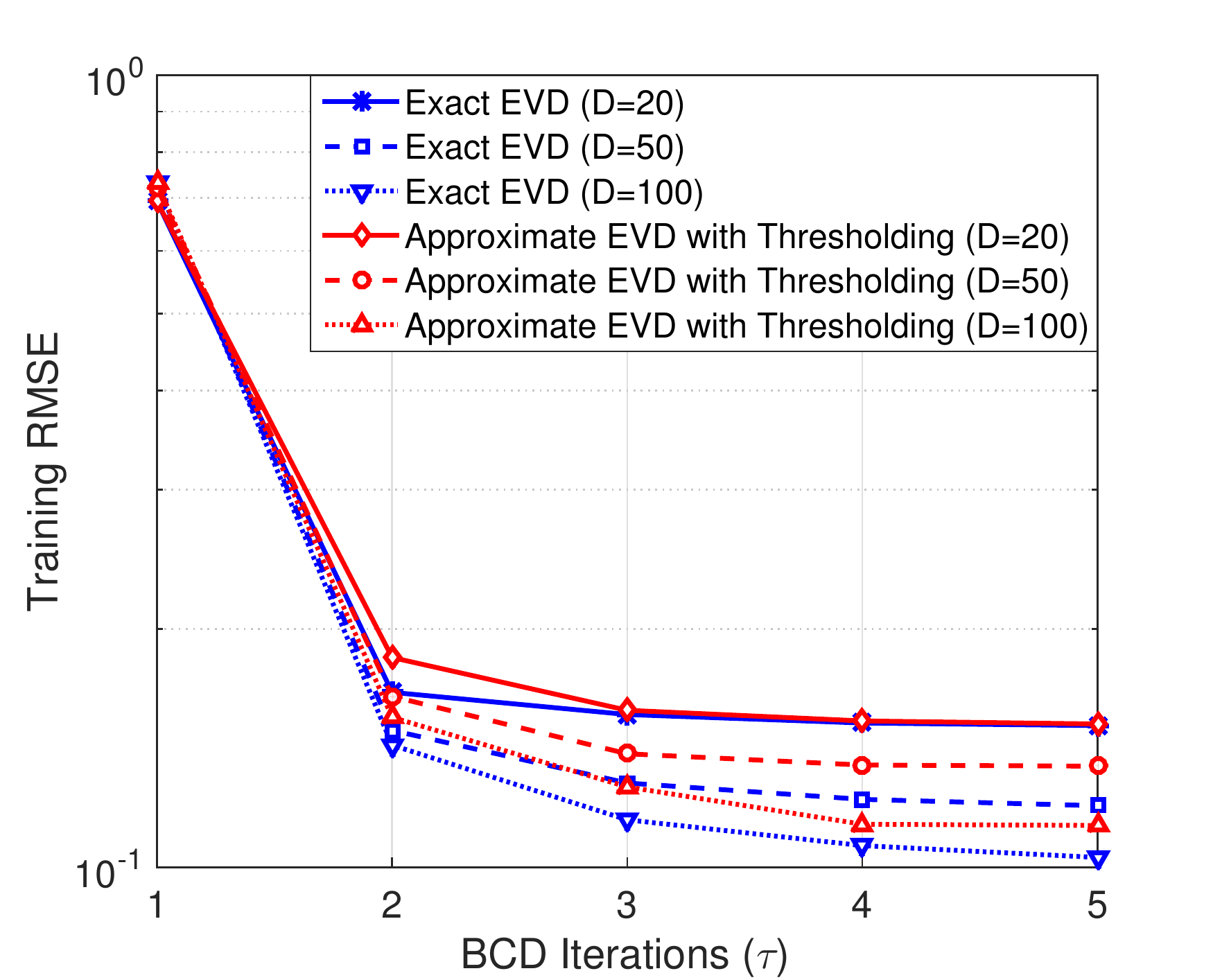}
}
\subfigure[]{
\includegraphics[width=7.0cm, trim=10 0 40 20, clip]{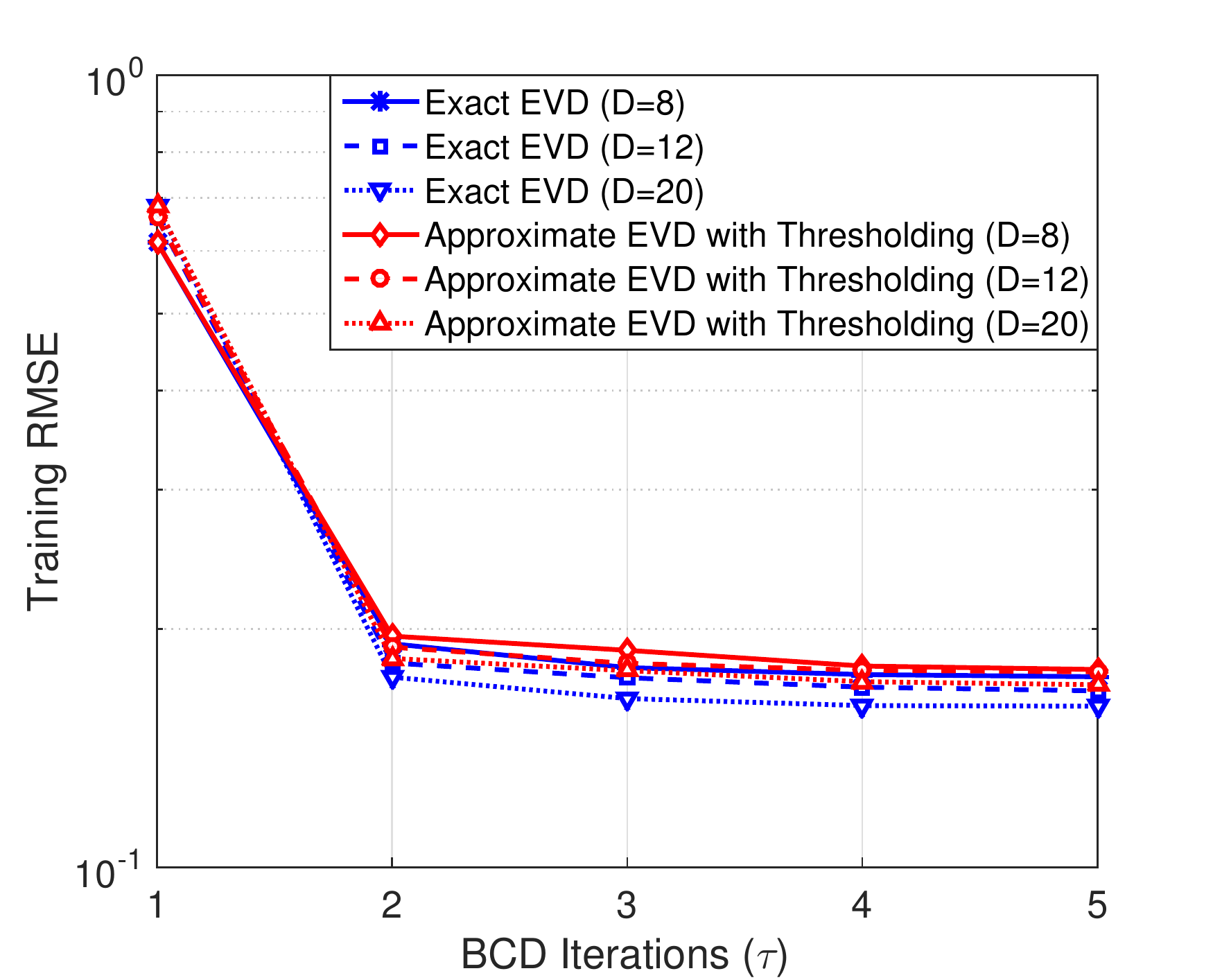}
}
\caption{Training RMSE vs. BCD iterations of the enhanced GD with exact EVD and approximate EVD with thresholding: (a) the artificial dataset (D1), (b) ML100K (D2), (c) ML1M (D3).} \label{Trainfig2}
\end{figure}

Next, we evaluate the performance of the enhanced GD with approximate EVD and thresholding technique presented in Section \ref{Sec.4.3} and compare it with Algorithm \ref{Algorithm: Enhanced Gradient Descent} (i.e., the enhanced GD with exact EVD) in Table \ref{Table:CC1} and Figs. \ref{Approximation Error}--\ref{Trainfig2}. The computational cost of the enhanced GD with approximate EVD and thresholding can be reduced significantly compared to that of exact EVD, especially when $D$ is large. It is worth noting that the time overhead of the enhanced GD with approximate EVD and thresholding does not increases substantially compared to the original GD (even the enhanced GD with exact EVD) as D grows from $D=20$ to $D=100$ for (D2). The same trend is observed for (D3). In Fig. \ref{Approximation Error}, we show the upper bounds of the approximation error of the approximate EVD with respect to $m$. It demonstrates that $\beta_{m+1}$ and $\beta_{m+1}|q_i(m)|$ in Lemma \ref{Lemma:Approximation error} provide tight upper bounds of $\max_i|\lambda_i-\hat{\lambda}_i|$ and $\min_i|\lambda_i-\hat{\lambda}_i|$, respectively. It also shows that the threshold value $\delta$ in \eqref{eq:threshold value}, based on Lemma \ref{Lemma:UB on beta}, guides a good choice of $m$. For instance of Fig. \ref{Approximation Error}, the approximate EVD terminates when $m=5$ (the point $\beta_{m+1}\leq \delta$) with the approximation error of the dominant eigenvalue far below $10^{-5}$. Moreover, as depicted in Fig. \ref{Trainfig2}, the training performance of the approximate EVD with thresholding is very close to that of the exact EVD.

\subsection{Prediction Performance}
\begin{table*}[t]
 \centering
 \caption{Time Consumption (in Seconds) Comparison of Solving the BQP} \label{Table:CC1}
 \begin{threeparttable}[b]
 \resizebox{\textwidth}{!}{
 \begin{tabular}{|c|c|c|c|c|c|c|c|}
  \hline
  Dataset & \diagbox{Algorithm}{$D$} & 4 & 8 & 12 & 20 & 50 & 100  \\
  \hline
  \multirowcell{6}{(D1)} & SDRwR\tnote{a} & $7.11$ & $7.36$ & $7.62$ & - & - & - \\
                         & DMO\tnote{a} & $1.21\times 10^{-1}$ & $1.85$ & $1.00\times 10^{+3}$ & - & - & - \\
                         & Quasi-Newton\tnote{b} & - & - & - & $1.91$ & $1.76\times 10^{+1}$ & $8.04\times 10^{+1}$ \\
                         & Original GD & $8.16\times 10^{-2}$ & $1.12\times 10^{-1}$ & $1.32\times 10^{-1}$ & $1.47$ & $1.48\times 10^{+1}$ & $6.76\times 10^{+1}$ \\
                         & Enhanced GD (with exact EVD) & $8.10\times 10^{-2}$ & $8.60\times 10^{-2}$ & $1.07\times 10^{-1}$ & $1.63\times 10^{-1}$ & $5.34\times 10^{-1}$ & $2.56$ \\
                         & Enhanced GD (with approximate EVD \& thresholding)\tnote{c} & - & - & - & $1.55\times 10^{-1}$ & $1.98\times 10^{-1}$ & $2.87\times 10^{-1}$ \\
  \hline
  \multirowcell{5}{(D2)} & SDRwR\tnote{a} & $2.99\times 10^{+2}$ & $3.08\times 10^{+2}$ & $3.18\times 10^{+2}$ & - & - & - \\
                         & DMO\tnote{a} & $2.02\times 10^{+1}$ & $7.80\times 10^{+1}$ & $7.57\times 10^{+3}$ & - & - & - \\
                         & Original GD & $2.14\times 10^{+1}$ & $2.24\times 10^{+1}$ & $2.41\times 10^{+1}$ & $1.02\times 10^{+2}$ & $5.53\times 10^{+2}$ & $5.10\times 10^{+3}$ \\
                         & Enhanced GD (with exact EVD) & $2.10\times 10^{+1}$ & $2.19\times 10^{+1}$ & $2.30\times 10^{+1}$ & $2.67\times 10^{+1}$ & $5.49\times 10^{+1}$ & $2.59\times 10^{+2}$ \\
                         & Enhanced GD (with approximate EVD \& thresholding)\tnote{c} & - & - & - & $2.58\times 10^{+1}$ & $3.20\times 10^{+1}$ & $3.87\times 10^{+1}$ \\
  \hline
  \multirowcell{2}{(D3)\tnote{d}} & Enhanced GD (with exact EVD) & - & $5.88\times 10^{+2}$ & $6.01\times 10^{+2}$ & $8.14\times 10^{+2}$ & - & - \\
                         & Enhanced GD (with approximate EVD \& thresholding) & - & $5.84\times 10^{+2}$ & $5.92\times 10^{+2}$ & $6.89\times 10^{+2}$ & - & - \\
  \hline
\end{tabular}}
    \begin{tablenotes}
    \footnotesize
    \item[a] The missed entries (`-') are due to the extraordinary high computational cost of the SDRwR and DMO when $D$ is large.
    \item[b] For the quasi-Newton method, we utilize (D1) and focus on the cases when $D$ is large.
    \item[c] The missed entries exist because we focus on evaluating the performance of the enhanced GD with approximate EVD \& thresholding when $D$ is large.
    \item[d] For the large-scale (D3), our focus has been switched to the proposed scalable enhanced GD and the simulations are done only for $D=8,12,20$.
    \end{tablenotes}
 \end{threeparttable}
\end{table*}

\begin{table*}[t]
 \centering
 \caption{Prediction Performance Evaluation (NRMSE Comparison) based on (D2) and (D3)} \label{Table:Test}
 \begin{threeparttable}
 \resizebox{\textwidth}{!}{
 \begin{tabular}{|c|c|c|c|c|c|}
  \hline
  Dataset & \diagbox{Algorithm}{$D$} & 4 & 8 & 16 & 24  \\
  \hline
  \multirowcell{8}{(D2)} & KM-Enhanced GD & 0.1978 & 0.1963 & 0.1946 & 0.1891 \\
                         & (Algorithm \ref{Algorithm: Overall KM learning})& $(\lambda_u=10,\mu_i=0)$  & $(\lambda_u=30,\mu_i=0)$  &$(\lambda_u=40,\mu_i=0)$  & $(\lambda_u=60,\mu_i=0)$ \\
                         \cline{2-6}
                         & KM-Enhanced GD & 0.2045 & 0.1999 & 0.1961 & 0.1914 \\
                         & with approximate EVD \& thresholding &$(\lambda_u=10,\mu_i=0)$ & $(\lambda_u=30,\mu_i=0)$ & $(\lambda_u=60,\mu_i=0)$ & $(\lambda_u=60,\mu_i=0)$ \\
                         \cline{2-6}
                         & \multirowcell{2}{NNM} & 0.1944 & 0.2255 & 0.2057 & 0.2118 \\
                         &                       & $(\lambda_u=10,\mu_i=0)$  & $(\lambda_u=20,\mu_i=0)$  &$(\lambda_u=40,\mu_i=0)$  & $(\lambda_u=50,\mu_i=10)$ \\
                         \cline{2-6}
                         & MF\tnote{d,e} & 0.2292 & 0.2287~$(k=10)$ & - & 0.2269~$(k=40)$ \\
                         \cline{2-6}
                         & SVD++\tnote{d} & 0.2284 & 0.2277~$(k=10)$ & 0.2270~$(k=20)$ & 0.2266~$(k=50)$ \\
  \hline
  \multirowcell{8}{(D3)} & KM-Enhanced GD & 0.1812 & 0.1768 & 0.1716 & 0.1629 \\
                         & (Algorithm \ref{Algorithm: Overall KM learning}) & $(\lambda_u=0,\mu_i=2)$  & $(\lambda_u=10,\mu_i=1.5)$  &$(\lambda_u=20,\mu_i=1.5)$  & $(\lambda_u=30,\mu_i=2)$ \\
                         \cline{2-6}
                         & KM-Enhanced GD & 0.2478 & 0.1812 & 0.1765 & 0.1684 \\
                         & with approximate EVD \& thresholding & $(\lambda_u=10,\mu_i=2)$ & $(\lambda_u=10,\mu_i=0.5)$ & $(\lambda_u=10,\mu_i=4)$ & $(\lambda_u=10,\mu_i=2.5)$ \\
                         \cline{2-6}
                         & \multirowcell{2}{NNM} & 0.1798 & 0.1776 & 0.1765 & 0.1758 \\
                         &                       & $(\lambda_u=0,\mu_i=1.5)$  & $(\lambda_u=10,\mu_i=2.5)$  &$(\lambda_u=10,\mu_i=1.5)$  & $(\lambda_u=10,\mu_i=3)$ \\
                         \cline{2-6}
                         & MF\tnote{d,e} &- &0.2143~$(k=10)$ &- &- \\
                         \cline{2-6}
                         & SVD++\tnote{d,e} &- &0.2130~$(k=10)$ &0.2128~$(k=20)$ &- \\
  \hline
\end{tabular}}
    \begin{tablenotes}
    \footnotesize
    \item[d] The results of MF and SVD++ are taken from the following repository: \textit{http://www.mymedialite.net/examples/datasets.html}.
    \item[e] The missed entries (`-') are due to the unavailability of the corresponding RMSE result of the repository.
    \end{tablenotes}
 \end{threeparttable}
\end{table*}

We assess the prediction performance of the proposed methods against the existing methods, including NNM \cite{Stark16}, MF \cite{Koren09}, and SVD++ \cite{Koren08}, based on the ML100K (D2) and the ML1M (D3) datasets. We adopt the normalized RMSE (NRMSE) as a metric, which is given by
\beq
E_{\text{test, NRMSE}} \!\triangleq \! \left\{\!\!\!
                             \begin{array}{ll}
                               \sqrt{\frac{1}{|\cT|}\sum_{(u,i)\in\cT}(p_{u,i} - {\bthe^\star_u}^T\bpsi^\star_i)^2}, \\
                                     \quad \quad \quad \quad \quad \quad \quad   \hbox{for KM and NNM} \\
                               \frac{1}{r_{\text{max}}-r_{\text{min}}}\sqrt{\frac{1}{|\cT|}\sum_{(u,i)\in\cT}(r_{u,i} - \hat{r}_{u,i})^2}, \\
                                     \quad \quad \quad \quad \quad \quad \quad   \hbox{for MF and SVD++}
                             \end{array},
                           \right. \nn
\eeq
where $r_{\text{max}}-r_{\text{min}}=5-1=4$ and $\hat{r}_{u,i}$ is the predicted rating score via MF or SVD++. The above normalization, which scales by $1/(r_{\text{max}}-r_{\text{min}})$ ensuring that the predicted values of all the different methods are contained in $[0,1]$, is widely used in ML \cite{Ghauch18}.
The NRMSE results of prediction on (D2) and (D3) are provided in Table \ref{Table:Test}. In Table \ref{Table:Test}, $\lambda_u$ and $\mu_i$ are two hyperparameters to mitigate overfitting by using cross validation \cite{Ghauch18}. Specifically, the value of $(\lambda_u, \mu_i)$ in each entry indicates the best parameter pair associated with corresponding method and $D$. To ensure a reasonable comparison, the size of factorization for MF and SVD++, i.e., $k$, is chosen to be as close as possible to $D$. It reveals that the KM with enhanced GD shows significantly better prediction performance compared to the benchmarks and the predication error gap between this method and the benchmarks improves with increasing $D$. This is attributed to the advantageous nature of KM that being an accurate model in a mathematical sense and rooted in probability theory, while other benchmarks are based on intuition.

\subsection{Interpretability via Logical Relation Mining} \label{Sec 5.3}
In Section \ref{Section 2.2.3}, we have briefly introduced the interpretability of KM. In order to exploit the logical relations between random variables $X_{u,i}$ and $X_{u,j}$ ($(u,i)\in\cK$ and $(u,j)\in\cK$) based on the optimized KM parameters $\bpsi_i^\star$ and $\bpsi_j^\star$, $\forall i,j\in\cI_{\cK}$, an indicator matrix of the logical relations, $\bN\in\B^{|\cI_{\cK}|\times |\cI_{\cK}|}$, also known as an adjacency matrix \cite{Ghauch18}, can be built as
\beq \label{adjacency matrix}
N(i,j) = \bigg\{
                                  \begin{array}{ll}
                                    1, & \hbox{if~supp$(\bpsi_j^\star)$ $\subseteq$ supp$(\bpsi_i^\star)$} \\
                                    0, & \hbox{\text{otherwise}}
                                  \end{array} ,\nn
\eeq
where the nonzero entry $N(i,j)=1$ shows that $X_{u,i}$ and $X_{u,j}$ are coupled and mutually influencial. Constructing $\bN$ allows us to further evaluate how much $X_{u,i}$ influences or is influenced by $X_{u,j}$, $\forall j\in\cI_{\cK}$, via introducing the normalized influence score \cite{Ghauch18} as
\beq \label{normalized influence score}
\varsigma_i = \frac{1}{|\cI_{\cK}|} \sum_{j\in\cI_{\cK}} N(i,j), ~\forall i\in\cI_{\cK}.
\eeq
Stated differently, $\varsigma_i$ counts the (normalized) number of relations that $X_{u,i}$ is logically connected to. In particular, $\varsigma_i= 1$ denotes a maximally supported random variable, i.e., $\bpsi_i^\star=\bones$ and $\text{supp}(\bpsi_j^\star)\subseteq \text{supp}(\bpsi_i^\star)$ holds $\forall j\in\cI_{\cK}$.

\begin{figure}[t]
\centering
\includegraphics[width=6.8cm, trim=20 05 50 30, clip]{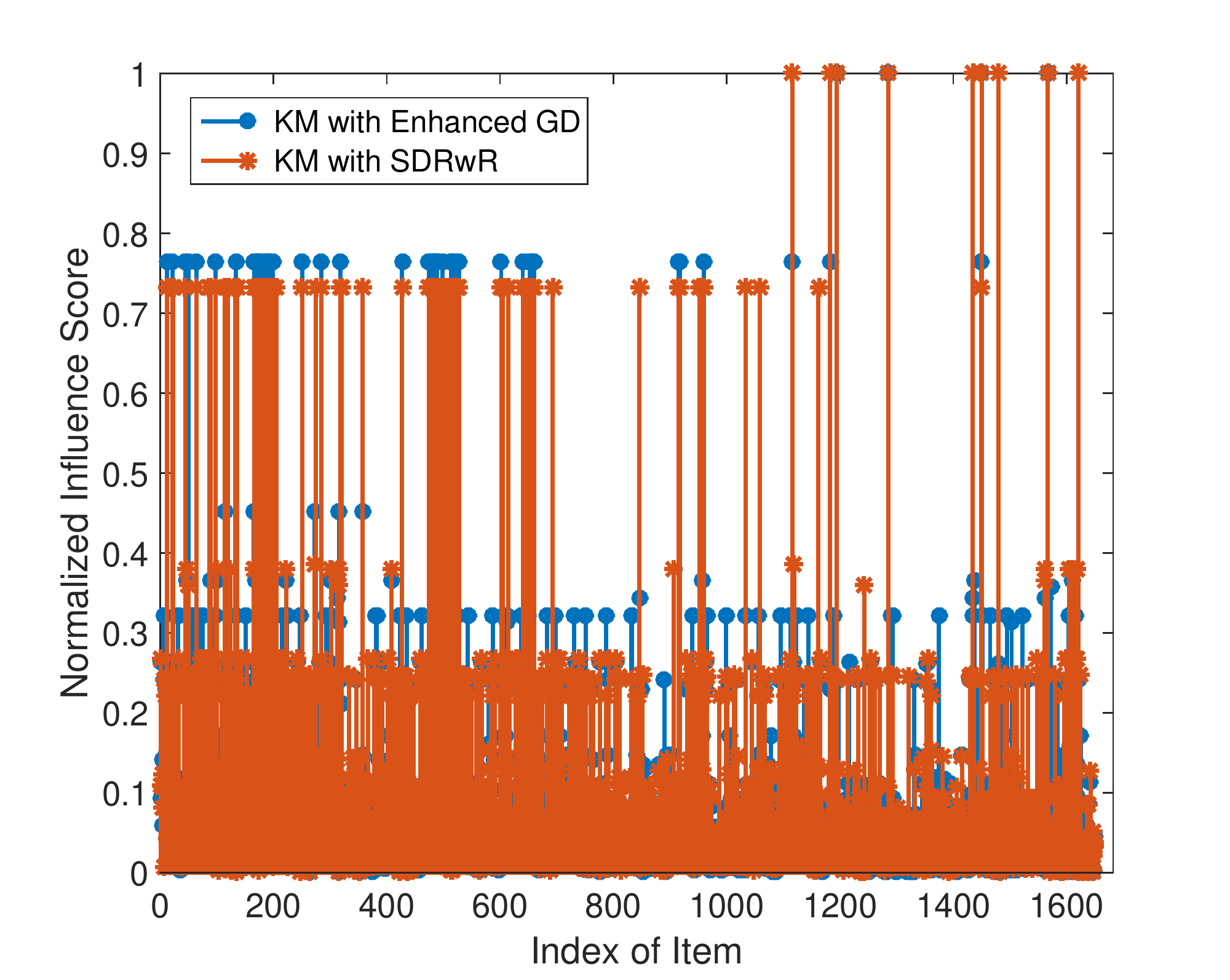}
\caption{Normalized influence score for two different algorithms on the ML100K dataset (D2) when $D=8$.} \label{CRMfig1}
\end{figure}

We display the normalized influence score, i.e., $\varsigma_i$ in \eqref{normalized influence score}, mined by two KM learning algorithms including the proposed KM with enhanced GD (Algorithm \ref{Algorithm: Overall KM learning}) and previous KM with SDRwR \cite[Algorithm 1]{Ghauch18}, for the ML100K dataset (D2). In Fig. \ref{CRMfig1}, we find that the results of logical relation mining are quite similar for the above two algorithms. However, the proposed KM with enhanced GD offers an order of magnitude reduction in the computational complexity, compared to the KM with SDRwR \cite{Ghauch20}. Furthermore, we confirm the efficacy of logical relation mining of Algorithm \ref{Algorithm: Overall KM learning} by identifying the set of items corresponding to $\varsigma_i=1$, as in Table \ref{Table:CRM}. Theoretically, if a user likes one of these items, then the user likes all other items in the training set. In Table \ref{Table:CRM}, the first column shows the item index with $\varsigma_i=1$, while the second column lists the user index in the training set who have rated the corresponding item. The total number of items rated by the user is shown in the third column. We calculate the accuracy of logical relation mining by setting a threshold to the empirical probability of the training set, i.e., $p_{u,i}\geq 50\%$, indicating that the user $u$ likes the item $i$.  For instance, the item of index 1201 has been rated by the user of index 90 and this user has rated 164 items in total. By checking the empirical probabilities of the ML100K dataset, we find that there are 153 items with $p_{u,i}\geq 50\%$. As observed from Table \ref{Table:CRM}, the accuracy of logical relation mining by using the KM with enhanced GD is above $80\%$.

\begin{table}[t]
 \centering
 \caption{Accuracy of logical relations mining for the ML100K dataset (D2).} \label{Table:CRM}
 \resizebox{\linewidth}{!}{
 \begin{tabular}{|c|c|c|c|}
  \hline
  \centering item index with $\varsigma_i=1$ & \centering user index & \centering \# of items rated & accuracy \\
  \hline
  \centering 1201 & \centering 90 & \centering 164 & 93.29\% \\
  \hline
  \multirowcell{3}{1293} & \centering 146 & \centering 19 & 84.21\% \\ & \centering 489 & \centering 109 & 80.73\% \\ & \centering 519 & \centering 49 & 81.63\% \\
  \hline
  \multirowcell{2}{1467} & \centering 244 & \centering 117 & 86.32\% \\ & \centering 886 & \centering 240 & 80.00\% \\
  \hline
  \centering 1599 & \centering 437 & \centering 238 & 82.77\% \\
  \hline
\end{tabular}}
\end{table}

\section{Conclusion}
In this paper, we presented a novel KM learning algorithm by using an enhanced GD approach based on dual optimization. To be specific, the BQP subproblem of KM learning was reformulated as a regularized dual optimization problem of strong convexity, which can be solved by GD. Considering the demand of scalability and the drawback of traditional GD due to a high reliance on the computation of EVD, we proposed an efficient enhanced GD with EVD elimination. Furthermore, a numerical approximate EVD was adopted to extract the spectra of symmetric matrices with low computational complexity. Inspired by the approximation error analysis, we explored the tractable bound which depends only on the traces and the normalized Minkowski $\ell_1$-norm, and then proposed a thresholding scheme for the approximate EVD. The proposed methods were applied to different datasets and numerical results demonstrated their superiority compared to other benchmarks in terms of computational cost, training/prediction performance, and interpretability.

\section*{Acknowledgement} 
We are deeply indebted to the reviewers and editor, whose consistent comments greatly improved the manuscript.

\appendices
\section{Proof of Lemma \ref{dual problem} } \label{Proof of Lemma 1}
\begin{proof}
  The Lagrangian of the primal problem in \eqref{regularized SDP} is given by
  \beq \label{Lagrangian}
  \!\!\!\!\!\!\!\!\!\!\!\!\!\!\!\! &&\cL(\bX,\bu,\bD) = \nn \\
  \!\!\!\!\!\!\!\!\!\!\!\!\!\!\!\! &&\langle\bX, \bA\rangle + \frac{1}{2\gamma}\|\bX\|_F^2 - \langle\bX, \bD\rangle + \sum_{i=1}^{D+1}u_i(\langle\bX, \bB_i\rangle - 1),
  \eeq
  where $\bu\in\R^{D+1}$ and $\bD\succeq\bzeros$ are Lagrangian multipliers. Since the problems in \eqref{regularized SDP} and \eqref{Lagrangian} are feasible, strong duality holds and $\nabla_{\bX}\cL(\bX^\star,\bu^\star,\bD^\star) = 0$, where $\bX^\star$, $\bu^\star$, and $\bD^\star$ are optimal solutions to \eqref{Lagrangian}. Then we have
  \beq \label{optimal solution of SDP relaxation-0}
  \bX^\star = \gamma\Big(\bD^\star - \bA - \sum_{i=1}^{D+1}u^\star_i\bB_i\Big) = \gamma(\bD^\star + \bC(\bu^\star)),
  \eeq
  where $\bC(\bu^\star)=-\bA-\sum_{i=1}^{D+1}u^\star_i\bB_i$.
  Substituting $\bX^\star$ in \eqref{Lagrangian}, we obtain the dual formulation
  \beq \label{dual formulation}
  \max_{\bu\in\R^{D+1},\bD\succeq\bzeros} -\bu^T\bones - \frac{\gamma}{2}\|\bD + \bC(\bu)\|_F^2.
  \eeq
  For a given $\bu$, the dual problem in \eqref{dual formulation} is equivalent to
  \beq \label{variable elimination}
  \min_{\bD\succeq\bzeros}\quad \frac{\gamma}{2}\|\bD + \bC(\bu)\|_F^2.
  \eeq
  The solution to \eqref{variable elimination} is $\bD^\star = \Pi_+(-\bC(\bu))$. Due to the fact that $\bC(\bu)=\Pi_+(\bC(\bu))-\Pi_+(-\bC(\bu))$, it follows
  \beq \label{eq:useful equality}
  \bD^\star+\bC(\bu)=\Pi_+(\bC(\bu)).
  \eeq
  Thus the dual formulation in \eqref{dual formulation} can be simplified to \eqref{eq:dual}.

 We take the first-order derivative of $d_\gamma(\bu)$ in \eqref{eq:dual} with respect to $\bu$ and obtain
 \beq
 \nabla_{\bu} d_\gamma(\bu) \!\!\!\!&=&\!\!\!\! -\bones - \gamma\nabla_{\bu}\Big(\frac{1}{2}\|\Pi_+(\bC(\bu))\|_F^2\Big) \nn \\
                            \!\!\!\!&=&\!\!\!\! -\bones + \gamma\Phi[\Pi_+(\bC(\bu))], \nonumber
 \eeq
 where the last equality is due to $\nabla_{\bU}(\frac{1}{2}\|\Pi_+(\bU)\|_F^2)=\nabla_{\bU}(\frac{1}{2}\sum_{i=1}^{N}(\max(0,\lambda_{\bU,i}))^2) = \Pi_+(\bU)$, where $\lambda_{\bU,i}$ is the $i$th eigenvalue of $\bU\in\R^{N\times N}$.
 This concludes the proof.
\end{proof}

\section{Proof of Lemma \ref{Lemma:Approximation error} } \label{Proof of Lemma 2}
\begin{proof}
  First, suppose the following decomposition
\beq
\bH &=& \bP^T\bC(\bu_i)\bP = [\bP_m, \bP_n]^T\bC(\bu_i)[\bP_m, \bP_n] \nn \\
    &=& \left[
          \begin{array}{cc}
            \bP_m^T\bC(\bu_i)\bP_m & \bP_m^T\bC(\bu_i)\bP_n \\
            \bP_n^T\bC(\bu_i)\bP_m & \bP_n^T\bC(\bu_i)\bP_n \\
          \end{array}
        \right]\nn \\
    &=& \left[
          \begin{array}{cc}
            \bH_m & \bH_{mn}^T \\
            \bH_{mn} & \bH_n \\
          \end{array}
        \right], \label{eq:matrix H}
\eeq
where $\bP\in\R^{(D+1)\times(D+1)}$ is an orthonormal matrix, $\bP_m\in\R^{(D+1)\times m}$ and $\bP_n\in\R^{(D+1)\times (D+1-m)}$ are two sub-matrices of $\bP$, and $\bH_{mn}\in\R^{(D+1-m)\times m}$ has only one nonzero entry on its top-right corner, i.e., $\bH_{mn}(1, m) = \beta_{m+1}$.
Given the above decomposition, we show the proofs of the upper bound on $r_e(\bC(\bu_i)\hat{\bv}_i,\hat{\lambda}_i\hat{\bv}_i)$, $\max_i |\lambda_i-\hat{\lambda}_i|$, and $\min_i |\lambda_i-\hat{\lambda}_i|$, respectively.

i) We compute
\beq
\|\bC(\bu_i)\hat{\bv}_i - \hat{\lambda}_i\hat{\bv}_i\|_2 \!\!\!\!&=&\!\!\!\! \|\bC(\bu_i)\bP_m\bq_i - \vartheta_i\bP_m\bq_i\|_2  \nn \\
                                                         \!\!\!\!&=&\!\!\!\! \|\bP^T\bC(\bu_i)\bP_m\bq_i - \vartheta_i\bP^T\bP_m\bq_i\|_2 \nn \\
                                                         \!\!\!\!&=&\!\!\!\! \left\| \left[
                                                                       \begin{array}{c}
                                                                         \bH_m\bq_i \\
                                                                         \bH_{mn}\bq_i \\
                                                                       \end{array}
                                                                     \right] -\left[
                                                                                \begin{array}{c}
                                                                                  \vartheta_i\bq_i \\
                                                                                  \bzeros \\
                                                                                \end{array}
                                                                              \right] \right\|_2 \nn \\
                                                         \!\!\!\!&\overset{(a)}{=}&\!\!\!\! \|\bH_{mn}\bq_i\|_2  \nn \\
                                                         \!\!\!\!&\overset{(b)}{=}&\!\!\!\! \beta_{m+1}|q_i(m)| \overset{(c)}{\leq} \beta_{m+1}, \label{2-norm of the approximation error}
\eeq
where $(a)$ follows from the fact that $\bH_m\bq_i=\vartheta_i\bq_i$, $(b)$ holds because of the special structure of $\bH_{mn}$ in \eqref{eq:matrix H}, and $(c)$ is due to the fact that $\bq_i$ is unit norm.

ii) Defining $\hat{\bH} \!\triangleq \! \Big[\!
                              \begin{array}{cc}
                                \bH_m & \bzeros \\
                                \bzeros & \bH_n \\
                              \end{array}
                           \!\Big]$ and $\tilde{\bH}\! \triangleq \!\Big[\!
                                                          \begin{array}{cc}
                                                            \bzeros & \bH_{mn}^T \\
                                                            \bH_{mn} & \bzeros \\
                                                          \end{array}
                                                        \!\Big]$, we have $\bH = \hat{\bH} + \tilde{\bH}$ and the eigenvalues of $\hat{\bH}$ include the eigenvalues of $\bH_m$, i.e., $\vartheta_1, \ldots, \vartheta_m$. Then, based on the perturbation theory \cite{Horn1991}, we obtain
\beq
|\lambda_i-\hat{\lambda}_i| \leq \|\tilde{\bH}\|_2 = \|\bH_{mn}\|_2 = \beta_{m+1}. \nn
\eeq

iii) Since $\hat{\bv}_i=(\bC(\bu_i)-\hat{\lambda}_i\bI)^{-1}(\bC(\bu_i)-\hat{\lambda}_i\bI)\hat{\bv}_i$ when $\hat{\lambda}_i \neq \lambda_i$, $\forall i$, the following holds
\beq \label{eq:inequality in Proof of Lemma 1-3}
1 = \|\bv_i\|_2 \leq \|(\bC(\bu_i)-\hat{\lambda}_i\bI)^{-1}\|_2\|\bC(\bu_i)\hat{\bv}_i-\hat{\lambda}_i\hat{\bv}_i\|_2.
\eeq
By assuming that $\bC(\bu_i) = \bV_{\bC}\bLam_{\bC}\bV_{\bC}^T$ where $\bLam_{\bC}=\text{diag}([\lambda_1,\cdots,\lambda_{D+1}]^T)$, we have
\beq \label{eq:equality in Proof of Lemma 1-3}
\|(\bC(\bu_i)-\hat{\lambda}_i\bI)^{-1}\|_2 \!\!\!\!&=&\!\!\!\! \|\bV_{\bC}(\bLam_{\bC}-\hat{\lambda}_i\bI)^{-1}\bV_{\bC}^T\|_2 \nn \\
                                           \!\!\!\!&=&\!\!\!\! \frac{1}{\min_i |\lambda_i - \hat{\lambda}_i|}.
\eeq
By substituting \eqref{eq:equality in Proof of Lemma 1-3} into \eqref{eq:inequality in Proof of Lemma 1-3}, we obtain
\beq
\min_i |\lambda_i - \hat{\lambda}_i| \leq r_e(\bC(\bu_i)\hat{\bv}_i, \hat{\lambda}_i\hat{\bv}_i) \!\!\!\!&=&\!\!\!\! \|\bC(\bu_i)\hat{\bv}_i-\hat{\lambda}_i\hat{\bv}_i\|_2 \nn \\
                                                                                                 \!\!\!\!&=&\!\!\!\! \beta_{m+1}|q_i(m)|, \nn
\eeq
where the last equality is due to \eqref{2-norm of the approximation error}.

This concludes the proof.
\end{proof}

\section{Proof of Lemma \ref{Lemma:UB on beta} } \label{Proof of Lemma 3}
\begin{proof}
  According to Algorithm \ref{Algorithm: Lanczos method}, we have
\beq
\beta_{j+1} \!\!\!\!&=&\!\!\!\! \|\bC(\bu_i)\bp_j - \beta_j\bp_{j-1} - \alpha_j\bp_j\|_2 \nn \\
            \!\!\!\!&\leq &\!\!\!\! \|\bC(\bu_i)\bp_j - \alpha_j\bp_j\|_2 + \beta_j, \nn
\eeq
where the inequality follows from the triangle inequality and the fact that $\bp_{j-1}$ is unit norm. Then,
\beq
\beta_{m+1} \!\!\!\!&\leq&\!\!\!\! \sum_{j=1}^{m}\|(\bC(\bu_i) - \alpha_j\bI)\bp_j\|_2 \nn \\
            \!\!\!\!&\leq&\!\!\!\! \sum_{j=1}^{m} \sigma_{\text{max}}(\bC(\bu_i) - \alpha_j\bI) \nn \\
            \!\!\!\!&\leq&\!\!\!\! \sum_{j=1}^{m} (\sigma_{\text{max}}(\bC(\bu_i)) + \sigma_{\text{max}}(\alpha_j\bI)) \nn \\
            \!\!\!\!&=&\!\!\!\! m\sigma_{\text{max}}(\bC(\bu_i)) + \sum_{j=1}^{m}|\alpha_j| \nn \\
            \!\!\!\!&\leq&\!\!\!\! 2m\sigma_{\text{max}}(\bC(\bu_i)),  \label{upper bound based on sigma}
\eeq
where the last inequality is due to $|\alpha_j| = |\bp_j^T\bC(\bu_i)\bp_j| \leq \sigma_{\text{max}}(\bC(\bu_i))$, $j=1,\ldots,m$.

By introducing a normalized Minkowski $\ell_1$-norm $\hat{\sigma}_{\text{max,Minkowski}}\!\!\triangleq \!\! \frac{1}{D+1}\!\sum_{\ell=1}^{D+1}\!\sum_{j=1}^{D+1}\!|C(\ell,j)|$ \cite{Lutkepohl1997, Park17}, which is an approximation of $\sigma_{\text{max}}(\bC(\bu_i))$, we obtain
\beq
\beta_{m+1} \!\!\!\!&\leq&\!\!\!\! 2m(\sigma_{\text{max}}(\bC(\bu_i)) - \hat{\sigma}_{\text{max,Minkowski}}) \!+\! 2m \hat{\sigma}_{\text{max,Minkowski}} \nn \\
            \!\!\!\!&\leq&\!\!\!\! 2m\big( (\sigma_{\text{max,UB}}-\sigma_{\text{max,LB}}) \!+\! \hat{\sigma}_{\text{max,Minkowski}} \big), \nn
\eeq
where the last inequality stems from the fact that $\sigma_{\text{max,LB}}\leq \sigma_{\text{max}}(\bC(\bu_i)) \leq \sigma_{\text{max,UB}}$ and $\sigma_{\text{max,LB}}\leq \hat{\sigma}_{\text{max,Minkowski}} \leq \sigma_{\text{max,UB}}$ \cite{Wolkowicz1980,Park17}.
This concludes the proof.
\end{proof}

%==============================================================================================

\bibliographystyle{IEEEtran}
\bibliography{IEEEabrv,Manuscript_KM}

\end{document}